\documentclass[letterpaper, 10 pt, journal, twoside]{IEEEtran}
\usepackage{amsmath,amsfonts}
\usepackage{algorithmic}
\usepackage{array}
\usepackage{textcomp}
\usepackage{stfloats}
\usepackage{url}
\usepackage{verbatim}
\usepackage{graphicx}

\usepackage{times}
\usepackage{epsfig}

\usepackage{amssymb}
\usepackage{caption}
\usepackage{subcaption}
\usepackage[dvipsnames]{xcolor}
\usepackage{dsfont}
\usepackage{multirow}
\usepackage{mathtools}
\usepackage{pifont}
\usepackage[export]{adjustbox}
\usepackage{hyperref}

\newcommand{\update}[1]{\textcolor{black}{#1}}

\newcommand{\ASN}{{\it ActiveStereoNet}}
\newcommand{\ACDC}{{\it ACDC-Net}}
\newcommand{\etal}{{\it et al. }}

\newcommand{\cmark}{\ding{51}}%
\newcommand{\xmark}{\ding{55}}%
\newcommand{\eg}{\textit{e}.\textit{g}.}
\newcommand{\ie}{\textit{i}.\textit{e}.}

\begin{document}
\title{Self-Supervised Depth Completion for Active Stereo}

\author{Frederik Warburg$^{1}$, Daniel Hernandez-Juarez$^{2}$, Juan Tarrio$^{2}$, Alexander Vakhitov$^{2}$, Ujwal Bonde$^{2}$, \\ Pablo F. Alcantarilla$^{2}$
\thanks{Manuscript received: September, 9, 2021; Revised December, 5, 2021; Accepted January, 3, 2022.}
\thanks{This paper was recommended for publication by Editor Cesar Cadena Lerma. Name upon evaluation of the Associate Editor and Reviewers' comments.} 
\thanks{The main part of this work was done while Frederik Warburg was an intern
at SLAMcore.}
\thanks{$^{1}$ Technical University of Denmark, \texttt{frwa@dtu.dk}, $^{2}$ SLAMcore,  \texttt{\{first.surname\}@slamcore.com}}
\thanks{Digital Object Identifier (DOI): see top of this page.}
}

\markboth{IEEE ROBOTICS AND AUTOMATION LETTERS. PREPRINT VERSION. ACCEPTED January, 2022}
{Warburg \MakeLowercase{\textit{et al.}}: Self-Supervised Depth Completion for Active Stereo}


\maketitle

\begin{abstract}

Active stereo systems are used in many robotic applications that require 3D information. These depth sensors, however, suffer from stereo artefacts and do not provide dense depth estimates.In this work, we present the first self-supervised depth completion method for active stereo systems that predicts accurate dense depth maps. Our system leverages a feature-based visual inertial SLAM system to produce motion estimates and accurate (but sparse) 3D landmarks. The 3D landmarks are used both as model input and as supervision during training. The motion estimates are used in our novel reconstruction loss that relies on a combination of passive and active stereo frames, resulting in significant improvements in textureless areas that are common in indoor environments. Due to the non-existence of publicly available active stereo datasets, we release a real dataset together with additional information for a publicly available synthetic dataset (TartanAir~\cite{Wang20iros}) needed for active depth completion and prediction. Through rigorous evaluations we show that our method outperforms state of the art on both datasets. Additionally we show how our method obtains more complete, and therefore safer, 3D maps when used in a robotic platform. 

\end{abstract}


\section{Introduction}\label{sec:intro}


\IEEEPARstart{D}{epth} sensors are revolutionising computer vision applications that require 3D information about the scene such as non-rigid 3D reconstruction, robotics and augmented reality~\cite{Lenz15ijrr,Newcombe15cvpr}. Active stereo~\cite{Nishikara84spie} in particular is an interesting depth sensor technology 
for robotics due to its compact design, small power consumption and lower costs. It extends passive stereo by projecting a textured pattern while internally employing hardware-accelerated classical stereo algorithms~\cite{Keselman17cvprw}. However, active stereo systems suffer from stereo artefacts such as edge flattening, occlusions and do not produce depth estimates for distant, reflective or dark surfaces (See Fig. \ref{fig:teaser}). 

\begin{figure*}
    \centering
    \includegraphics[page=1,width=0.9\textwidth]{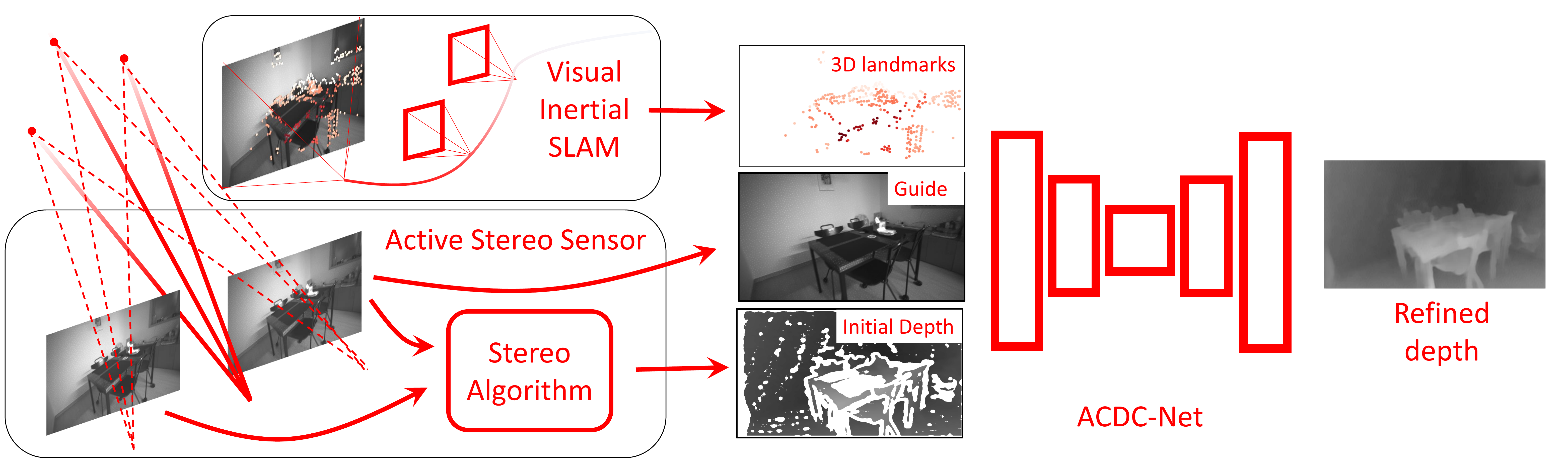}
    \caption{
    We propose a depth completion method to improve the accuracy of active stereo depth maps with the help of a visual-inertial SLAM system. The active stereo sensor has an IR stereo pair and a laser projector. We use the left IR image, the depth map from the sensor and the projections of the sparse 3D landmarks tracked by the visual-inertial SLAM system as input to the proposed \ACDC. The network outputs a completed and refined depth map in real-time.}
    \vspace{-0.5cm}
    \label{fig:overview}
\end{figure*}

\ASN~\cite{Zhang18eccv} is a recent approach developed for active stereo systems. They propose a learnable stereo pipeline for active stereo systems using rectified IR images with the visible active pattern. Their method predicts a dense depth map but suffers in well-lit rooms where the intensity of the active pattern is low.
While \ASN~uses photometric consistency between the stereo rectified IR images, we make one step further and leverage temporal consistency between estimated depth maps. It is important to note that this is a non-trivial extension since the projector moves alongside the camera. For this purpose, we work with the active stereo system in an interleaved fashion (one stereo frame with active illumination followed by one without) and use accurate 6-\textit{degrees of freedom} (DoF) trajectories from a feature-based visual-inertial Simultaneous Localisation and Mapping (SLAM) system inspired by~\cite{Leutenegger15ijrr}. Fig.~\ref{fig:overview} depicts an overview of our method: The visual-inertial SLAM operates on the stereo IR frames without active illumination and IMU measurements (gyroscope and accelerometer), while our network completes the semi-dense depth maps obtained from the frames with the active illumination. As an extra benefit of the closely integrated SLAM system, we show that using the sparse 3D landmarks that are tracked and refined over multiple views as an extra input and source of supervision improves the final output. We benchmark our ACtive Depth Completion-Net~(\ACDC) on diverse synthetic and real data for indoor scenes showing improved performance compared to state of the art self-supervised depth completion, depth from stereo and traditional dense stereo matching methods. 
We also show that our method improves the down-steam task of 3D mapping for robotics applications. 
Our main contributions can be summarised as: 

\begin{figure}
\centering
\includegraphics[width=.45\textwidth]{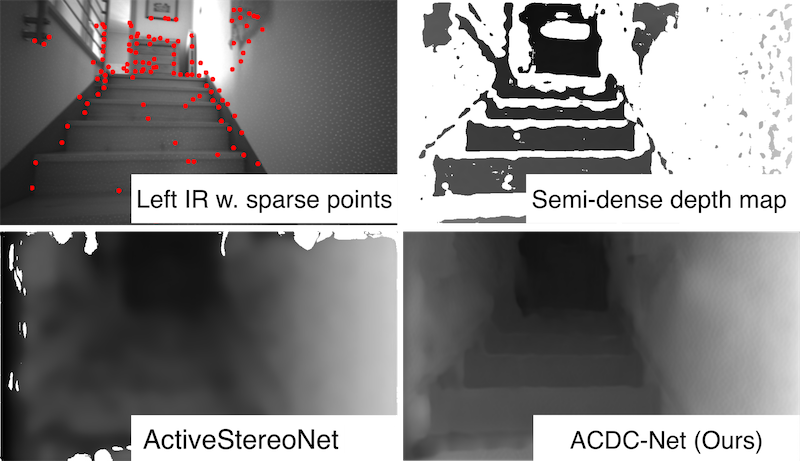}
\caption{We complete depth maps from an active stereo sensor with guidance from aligned IR frames and sparse 3D landmarks. Our proposed method \ACDC~produces high quality depth maps compared to state of the art~\cite{Zhang18eccv}.}
\vspace{-0.3cm}
\label{fig:teaser}
\end{figure}

\begin{itemize}
\itemsep0em 
    \item We propose the first end-to-end self-supervised active depth completion method. Rather than predicting a depth map from scratch, we complete the built-in depth prediction with the guidance of an aligned IR image.
    \item Our novel, self-supervised loss relies on a combination of warped active and passive frames, producing stronger signal in texture-less areas compared to the photometric loss on passive images only.
    \item We show how sparse 3D landmarks from a SLAM system can be used as an extra input and supervision to improve depth completion with a channel exchange network~\cite{Wang20arxiv}.
    \item Due to the non-existence of active stereo datasets, we release (upon paper acceptance) synthetic and real world datasets for active stereo depth completion and prediction. 
\end{itemize}

\vspace{-0.1cm}
\section{Related Work}\label{sec:related}
\vspace{-0.1cm}

\textbf{Learning-based active stereo} has had limited research in recent years. Prior to the deep learning era, frameworks for learning embeddings where matching can be performed more efficiently were explored~\cite{Fanello17iccv} together with direct mapping from pixel intensities to depth~\cite{Fanello14tog}. These methods have failed in general textureless scenes due to shallow architectures and local optimization schemes. More recently, \ASN~\cite{Zhang18eccv} proposed a self-supervised stereo method that estimates depth directly from the infrared (IR) images with active illumination. Their main contribution lies in the new weighted local contrast normalisation loss, which extends the photometric loss to handle images with active illumination. However, we find that this loss experiences convergence issues if the active pattern is not strong enough, which is often the case in well-lit rooms and outdoor environments. Similarly, \cite{Riegler19cvpr} proposes a learnable pipeline for monocular depth prediction with active illumination, which uses geometric and disparity losses in addition to photometric consistency \update{between an image and a reference pattern}. In contrast to these two methods, we hypothesize that depth completion can be faster and more reliable than depth estimation. 

Learning-based depth completion methods can be divided into \textbf{supervised}~\cite{Ma17icra,Senushkin20arxiv} and \textbf{self-supervised} methods~\cite{Godard19iccv,Tiwari20eccv,Wong20icra}. In this work, we will focus on self-supervised methods. Self-supervised depth completion and depth prediction methods rely on the \textit{photometric} loss~\cite{Zhou17cvpr}, that measures the difference in image intensities between a reference image and an image warped into this reference frame. However, this loss is problematic in poorly textured areas due to the lack of consistent geometry and it does not work well for spatially distant frames due to the large changes in appearance. To address these shortcomings, some works such as~\cite{Zhan19icra} introduces multi-view photometric and depth-normal consistency during training for depth prediction problems. Some recent batch approaches like~\cite{Tiwari20eccv} make use of Structure from Motion~(SfM) or visual SLAM frameworks to produce dense and geometrically consistent depth estimates for monocular videos. Sinha \etal~\cite{Sinha20eccv} proposed a supervised approach for depth prediction that first predicts a sparse set of 3D points from multiple views that are later densified by an encoder-decoder architecture that fuses depth and image features. In this work, we leverage a feature-based visual-inertial SLAM system to produce motion estimates and accurate (but sparse) 3D landmarks to self-supervise a convolutional neural network for depth completion for active stereo systems.
\textbf{Learning-based depth completion} has shown to attain a higher level of robustness and accuracy compared to monocular depth prediction, which is inherently ambiguous and unreliable~\cite{Ma19icra}. Most work has focused on completing sparse LiDAR images either considering guidance from RGB images~\cite{Eldesokey20pami,Ma19icra,Ma17icra} or without any guidance~\cite{Eldesokey20cvpr,Uhrig17ic3dv}. Uhrig \etal~\cite{Uhrig17ic3dv} found that sparse convolutions that explicitly consider the sparse nature of LiDAR data performs better than standard convolutions. Eldesokey \etal expand on these findings and show that normalised convolutions can propagate binary~\cite{Eldesokey20pami} or continuous learned~\cite{Eldesokey20cvpr} confidence through the network. Another line of depth completion methods tries to densify sparse depth prior to feeding it as input to a refinement network. \cite{Wong20icra} use scaffolding to densify sparse depth for visual odometry system, whereas \cite{chen18icvpr} use nearest neighbor upsampling and a heatmap of the locations of the sparse landmarks. These methods are well-suited for sparse LiDAR data that is rather uniformly distributed across the image but add little value to the semi-dense depth predictions of active stereo systems, which are dense in some regions but have large holes in other regions.

\vspace{-0.1cm}
\section{Method}\label{sec:method}
\vspace{-0.1cm}

\begin{figure*}
\vspace{0.2cm}
    \centering
        \includegraphics[page=2,width=0.9\textwidth]{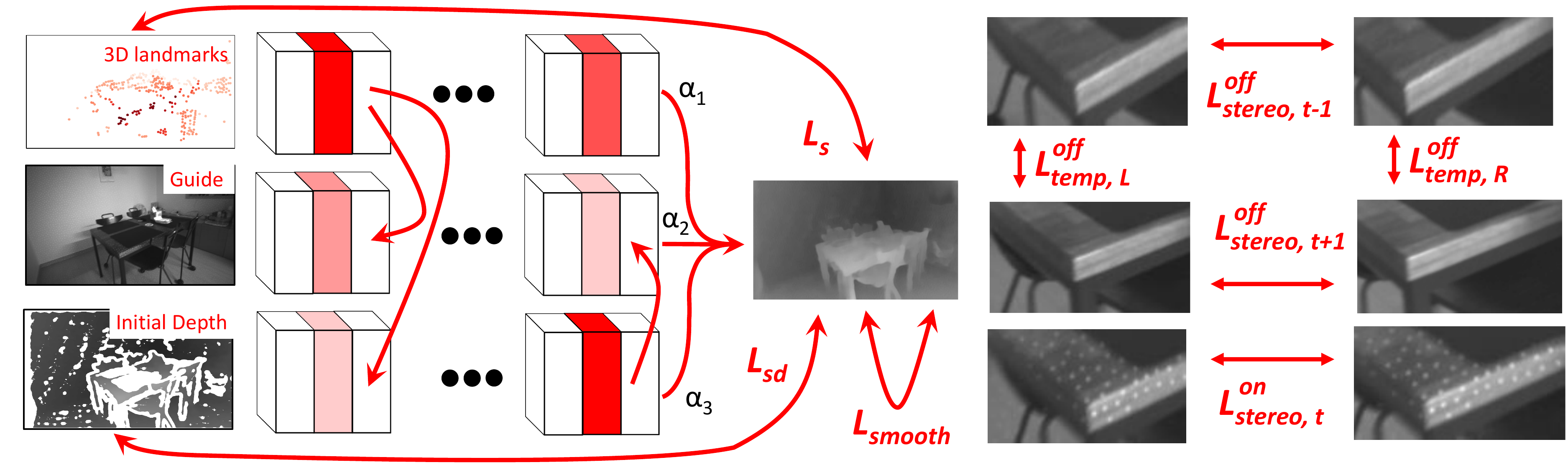}
    \caption{\ACDC~processes the initial depth map, guide image and rasterised 3D landmarks using three instances of the same backbone with shared weights, except for the batch normalization (BN) layers. It uses 'channel exchange' mechanism~\cite{Wang20arxiv}, replacing feature maps with low importance with most important ones from other network instances, estimating importance as BN scaling magnitude. During training we use the output depth map to compute photometric losses $L_{stereo}^{on}$,$L_{stereo,\{t-1, t+1\}}^{off}$, $L_{temp,\{L,R\}}^{off}$,
    consistency losses for the initial depth ($L_{sd}$) and sparse landmark projections ($L_s$), a smoothness loss  ($L_{smooth}$), and a sparsity-encouraging loss $L_{\gamma}$ for the BN scaling factors (not shown).
    }\vspace{-0.3cm}
    \label{fig:losses}
\end{figure*}


The main goal of our method is to complete and refine depth maps produced by an active stereo system, see Fig.~\ref{fig:overview}. To do that efficiently, we leverage a  feature-based  visual  inertial SLAM system to produce motion estimates and accurate (but sparse) 3D landmarks for supervision of the depth completion network. In particular, we propose an end-to-end pipeline that takes the following inputs: i. a reference IR frame captured by the active stereo sensor; ii. a semi-dense depth map produced by the active stereo sensor using a classical stereo-method and iii. an image where pixels are either zero or equal to the depth of the 3D landmarks tracked by the SLAM system and projected to the reference IR frame. Our method outputs a depth map aligned with the reference IR frame and works in real-time on a modern GPU. In our system, we run stereo visual-inertial SLAM on IR images without active illumination, obtained from the same sensor in the so-called {\em interleaved} mode, but our solution can be used alongside other types of SLAM and VO (\eg monocular RGB-D SLAM). 


\vspace{-0.1cm}
\subsection{ACDC-Net} 
\vspace{-0.1cm}

Our network takes an initial depth map, a guidance image and sparse 3D landmarks projected into the current frame as inputs, and outputs a refined depth map, see Fig.~\ref{fig:losses}. Previous depth completion methods have combined multi-modal input with either aggregation based methods such as concatenation and addition~\cite{Chen19arxiv,Eldesokey20pami} or alignment based methods~\cite{zhang18cvpr}. However, these methods are inadequate to balance the trade-off between inter-modal fusion and intra-model processing. To address this, we adapt the channel exchange framework~\cite{Wang20arxiv} that dynamically and parameter-free performs multi-modal fusion by exchanging feature channels between $M$ identical sub-networks~$f_m$. The channel exchange is self-guided by individual channel importance that is measured by the magnitude of Batch-Normalisation (BN) scaling factor $\gamma_{m, l}$ during training. The final output is a learned linear combination of the sub-networks. Since the sub-networks share all weights, the ensemble can be computed efficiently in parallel on a GPU.

Each sub-network is equipped with BN layers containing the scaling factors $\gamma_{m,l}$ for the $l^{th}$ layer. A channel is replaced if the magnitude of the sub-network-specific BN scaling factor $\gamma_{m,l}$ is below a fixed threshold.
As opposed to \cite{Wang20arxiv} that replaces with average channel signal across the sub-networks, we propose to replace with the channel with the strongest signal, \ie~$\max$ over the sub-networks. 

\vspace{-0.1cm}
\subsection{Visual Inertial SLAM}\label{sec:vslam}
\vspace{-0.1cm}

Our visual inertial SLAM system is inspired by the approach presented in~\cite{Leutenegger15ijrr}. We combine reprojection and IMU errors and formulate the multi-sensor fusion SLAM problem as a factor graph, where the goal is to estimate a set of navigation states (camera poses, speed and biases) plus a collection of 3D landmarks given a set of measurements that includes visual and IMU measurements. For simplicity, we assume that the intrinsics and extrinsics calibration of the sensor rig are given. Note that the visual inertial SLAM operates only on the stereo frames without active illumination. 

For a particular stereo frame in time with active illumination, we query from the SLAM system the tracked landmarks from the previous frame without active illumination and project those into the active stereo frame using the estimates of the 3D landmarks, sensor pose and known intrinsics and extrinsics. In this way we generate a depth image where pixels are either zero or equal to the depth of the projected 3D landmarks. 
We can get an estimate of the sensor pose for frames with active illumination by integrating IMU measurements between two consecutive states~\cite{Forster17tro}. \update{Even though our SLAM system operates in real-time, the poses and 3D landmarks used for supervision during training come from a batch bundle adjustment optimisation step that uses the passive frames, IMU measurements and loop closures during the sequence}.

Our SLAM system uses the A-KAZE~\cite{Alcantarilla13bmvc} feature extractor, although the method is general enough to use other types of handcrafted or deep learning-based feature extractors.

\vspace{-0.1cm}
\subsection{Training} 
\vspace{-0.1cm}

Fig.~\ref{fig:losses} depicts an overview of the losses used in our method. 

\noindent\textbf{Photometric Loss Review.} Using an obtained pose $T_{T \rightarrow S}$ between a source view $I_{S}$ and a target view $I_{T}$, we seek to predict the dense depth map $\hat{D}_T$ that minimises the photometric reprojection error $L_p$~\cite{Godard19iccv}:

\begin{align}
     L_{p}(I_T, I_{S \rightarrow T}) &= pe( I_T, I_{S \rightarrow T} ) \\
    \text{and } \quad I_{T \rightarrow S} &= I_S \langle proj(\hat{D}_T, T_{T \rightarrow S}, K) \rangle \label{eq:sample},
\end{align}
where $pe()$ is a functional that operates over the whole image,
$proj()$ maps $\hat{D}_T$ into 2D coordinates in frame $S$, given $T_{T \rightarrow S}$ and camera extrinsics $K$, and $\langle \rangle$ is the sampling operator. We use bilinear sampling to query from the source images~\cite{Godard19iccv,Zhou17cvpr} . We follow common practice~\cite{Godard19iccv} and define $pe()$ as a weighted average of SSIM and $l_1$ norm with weights~$0.85$~and~$0.15$. \update{The photometric loss assumes a static scene. To obey this assumption, we use a small temporal baseline and recordings in mostly static environments.}

\noindent\textbf{Photometric Loss for Interleaved Mode.} The loss must satisfy the photometric consistency constraint, which assumes a static Lambertian scene with constant illumination between source and target frame. This constraint holds between active stereo images, but not between active and passive frames or even between consecutive active frames ($I_{t}^{on}$ and $I_{t+2}^{on}$) as the projector moves with the camera ($t$ indexes over time). 
Similar to \ASN~we compute the photometric reprojection error between the stereo frames with the projector on $L_{stereo}^{on}$.

However, we take one step further by enforcing temporal consistency. We reproject both the previous and next passive images into the current view at time $t$. These passive images satisfy the photometric consistency constraint as neither have the active pattern. We compute the temporal losses, $L_{temp, R}^{off}$ and $L_{temp, L}^{off}$, for both the left/right stereo frame \update{in the current left view. Thus, all photometric losses are computed in the current left view using the depth of this frame for reprojection.} 

Lastly, 
we compute the passive stereo losses, $L_{stereo, t-1}^{off}$ and $L_{stereo, t+1}^{off}$, for the next and previous frames. 

\begin{figure}
    \vspace{0.2cm}
     \centering
     \begin{subfigure}[b]{0.45\linewidth}
         \centering
         \includegraphics[width=\textwidth]{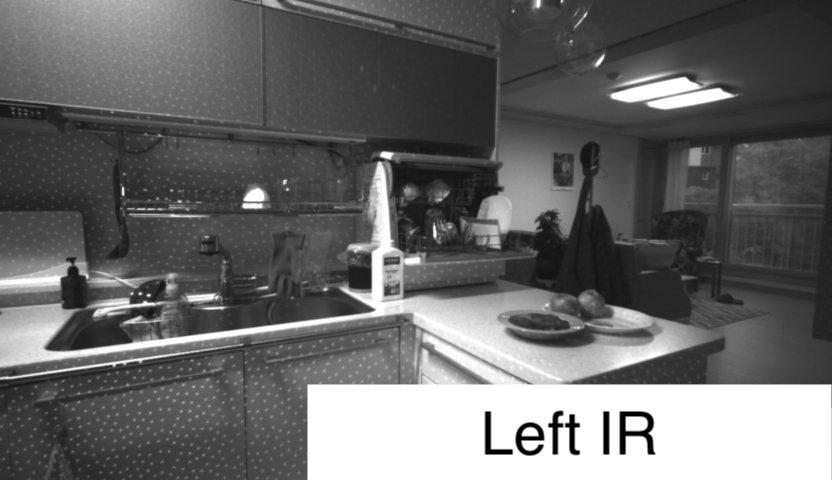}
     \end{subfigure}
     \begin{subfigure}[b]{0.45\linewidth}
         \centering
         \includegraphics[width=\textwidth]{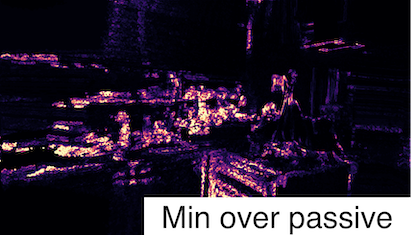}
     \end{subfigure} \\ \vspace{0.01cm} 
     \begin{subfigure}[b]{0.45\linewidth}
         \centering
         \includegraphics[width=\textwidth]{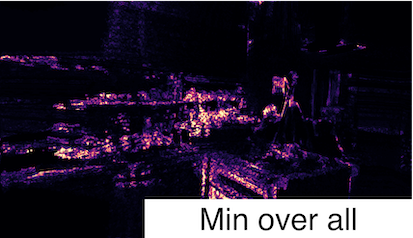}
    \end{subfigure}
     \begin{subfigure}[b]{0.45 \linewidth}
         \centering
         \includegraphics[width=\linewidth]{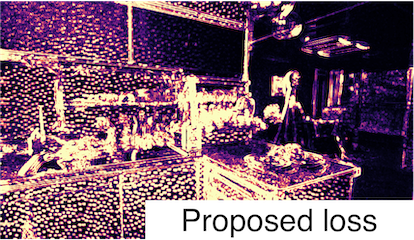}
     \end{subfigure}
        \caption{Applying the minimum technique~\cite{Godard19iccv} over passive and active losses kills all signal from the active loss. Therefore, we propose to take the $\min$ only over the temporal, passive losses. This results in a strong signal on textureless surfaces. \update{Lighter color means higher signal.}}
        \label{fig:passive_active_loss}
        \vspace{-0.6cm}
\end{figure}

\noindent\textbf{min-Operator for Interleave Mode.} 
\update{When computing the photometric error from multiple views, occluded pixels that are only visible in the target image can cause problems. If the network predicts the correct depth, the corresponding color in an occluded source image will likely not match the target, which leads to a high photometric error. Godard \etal\cite{Godard19iccv} introduced a $\min$-operator between two temporal reconstruction losses to remove occluded pixels from the loss, however applying this solution across all losses removes all signal from the active pattern as seen in Fig.~\ref{fig:passive_active_loss}.} We propose to split the $\min$ operation into an active and passive part. \update{Fig.~\ref{fig:passive_active_loss} shows that the proposed split} preserves signal in the texture-less areas. The occlusion artefacts mainly stem from the temporal losses as the baseline between temporal frames is larger than between the stereo frames. Therefore, it is enough to perform the $\min$ operation over the temporal frames.

\begin{align}\label{eq:photo_temporal}
    L_{photo} &= L_{stereo}^{on}+ \beta \min(L_{photo}^{off}),\\
    L_{photo}^{off} &= \left\{ L_{temp, R}^{off}, L_{temp,L}^{off}, L_{stereo, t-1}^{off}, L_{stereo,t+1}^{off} \right\},\nonumber
\end{align}
where we set $\beta = 1$. 

To filter out stationary pixels, we apply auto-masking~\cite{Godard19iccv}. 
We obtain the auto-mask by comparing the photometric projection error from source to target image with the identity reprojection. 
We apply this auto-masking for all passive losses.

\noindent\textbf{Input-output losses.} For large parts of the scene, the active stereo sensor computes reliable semi-dense depth estimates $D_{sd}$. We use the $l_1$-norm to penalise the difference between~$D_{sd}$ and our depth prediction $\hat{D}$ over $\Omega_{sd}$, where semi-dense depth is available.

\begin{equation}
    L_{sd} = \sum_{i \in \Omega_{sd}} w(i) \| D_{sd}(i) - \hat{D(i)} \|_1.
\end{equation}



\update{where $w(i) = 1/D_{sd}(i)^2$ such that distant regions are penalized less by the input-output loss.} Also, we use the depth of the sparse points from the SLAM system as supervision:
\begin{equation}\label{eq:sparse_loss}
    L_{s} = \sum_{i \in \Omega_s} \| D_{s}(i) - \hat{D}(i) \|_1.
\end{equation}

These features are more sparse than the semi-dense depth estimates, but more reliable, as they are tracked and refined over several frames. The sparse features are located at boundaries, and at distant objects, where the semi-dense depth maps are often missing depth measurements due to occlusions or the longer range. Therefore, we find that these two losses complement each other and speed up training significantly.

\noindent\textbf{Smoothness loss.} This loss is necessary to improve convergence and avoid the aperture problem. As our guidance image has the active pattern, we cannot directly apply the standard edge-guided smoothness loss \cite{Godard19iccv}. Instead, we use a $9 \times 9$ median filter on the infrared frame to remove the projected pattern followed by an edge-guided smoothness loss:


\begin{equation}
    L_{smooth}=\left|\partial_{x} d_{t}^{*}\right| e^{-\left|\partial_{x} I_{t}\right|}+\left|\partial_{y} d_{t}^{*}\right| e^{-\left|\partial_{y} I_{t}\right|}.
\end{equation}



\noindent\textbf{Sparsity regularisation.} Following \cite{Wang20arxiv}, we add an $l_1$ regularisation to the BN scaling factors $\gamma_{m,l}$ to encourage channel exchange between the different sub-networks:

\begin{equation}\label{eq:gamma_regularization}
    L_{\gamma} = \sum_m \sum_l \| \gamma_{m, l} \|_1,
\end{equation}
where $m$ indexes over sub-networks and $l$ over BN layers.

\noindent\textbf{Final loss.} Combining these losses gives us the final loss:

\begin{equation}
    L = \sum_l \frac{w_1}{l^2} L^l_{photo} + \frac{w_2}{l^2} L^l_{sd} + \frac{w_3}{l^2} L^l_{s} + \frac{w_4}{l^2} L^l_{sm}  + w_5 L_{\gamma},
\end{equation}
where $l$ indexes over several scales, and we set $w_1 = 1,~w_2 = 0.01,~w_3 = 1,~w_4 = 10^{-5},~w_5 = 2 \times 10^{-6}$. We compute the losses at four scales. We summarise these losses in Fig. \ref{fig:losses}. Note that all losses are averaged over valid pixels. 

\vspace{-0.1cm}
\section{Experimental Results}\label{sec:results}
\vspace{-0.1cm}

\vspace{-0.1cm}
As there does not exist any public available depth completion or prediction datasets for active stereo sensors, we curate and present two new datasets: a synthetic dataset and a real-world dataset. In this section, we describe these datasets. 

\vspace{-0.2cm}
\subsection{Datasets}\label{sec:dataset}
\vspace{-0.1cm}


\noindent \textbf{Active TartanAir} extends the TartanAir dataset~\cite{Wang20iros} for active stereo systems with realistic semi-dense depth maps computed using Semi-global Matching (SGM)~\cite{Hirschmuller08pami} and simulated IMU measurements. 
We simulate an active stereo sensor by rendering a textured pattern into the scene using the ground truth depth. The pattern was obtained by analysing the pattern projection on a white wall from the D435i sensor. This way we could detect blob positions by using a Difference of Gaussians filter together with Non-Maximum Suppression. Recording in interleaved mode was also used to increase contrast by computing the difference between projector on and off. 

\begin{figure}[t]
\vspace{0.2cm}
\centering
\begin{subfigure}{.45\linewidth}
  \centering
  \includegraphics[width=\linewidth]{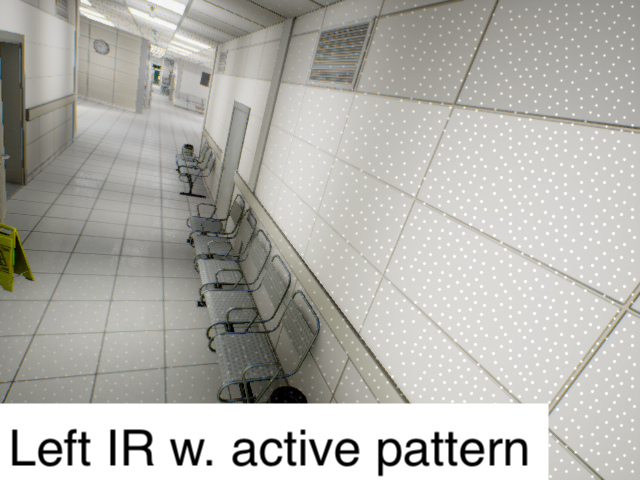}
\end{subfigure}%
\begin{subfigure}{.45\linewidth}
  \centering
  \includegraphics[width=\linewidth]{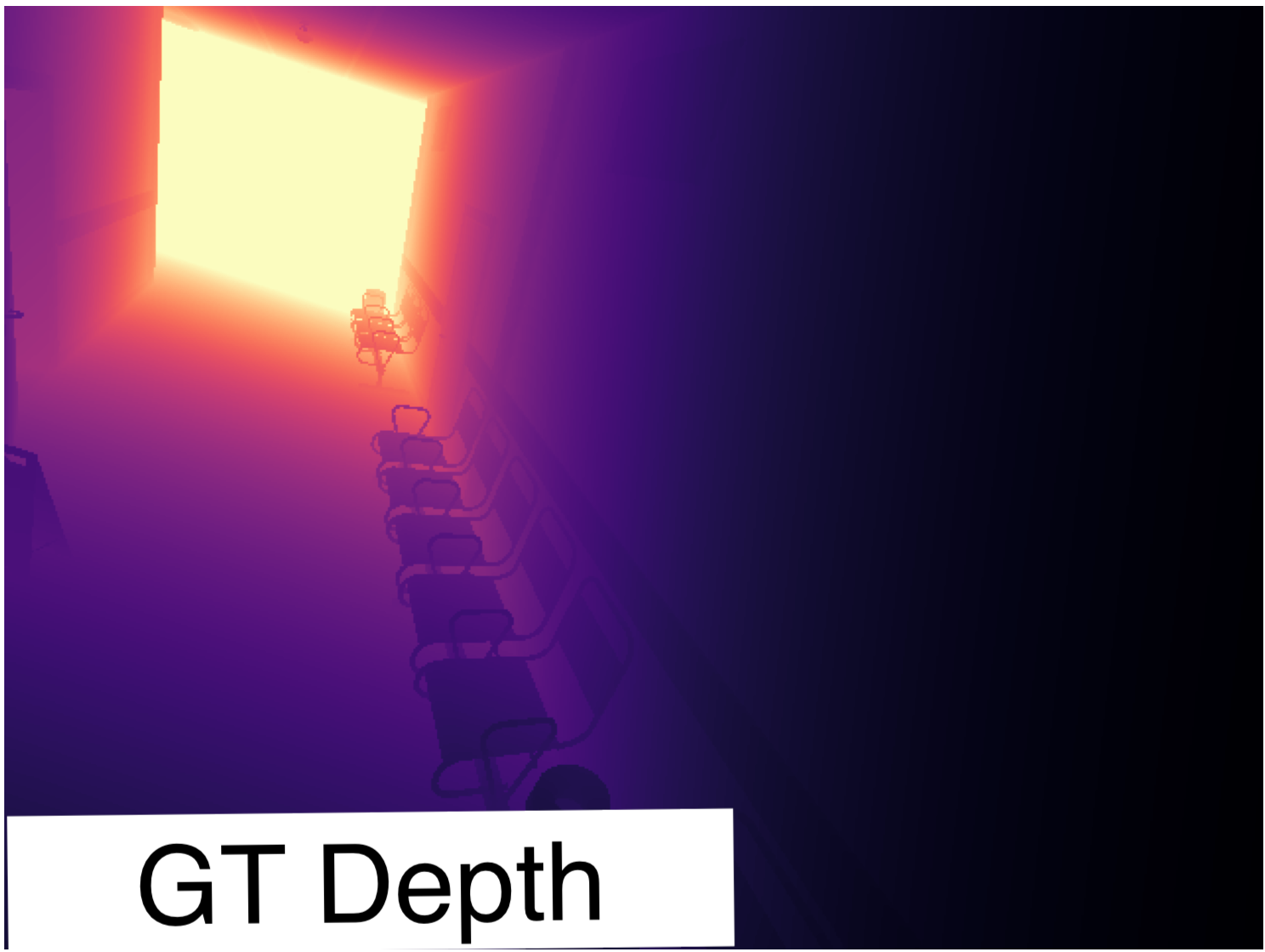}
\end{subfigure}\\ \vspace{0.01cm} 
\begin{subfigure}{.45\linewidth}
  \centering
  \includegraphics[width=\linewidth]{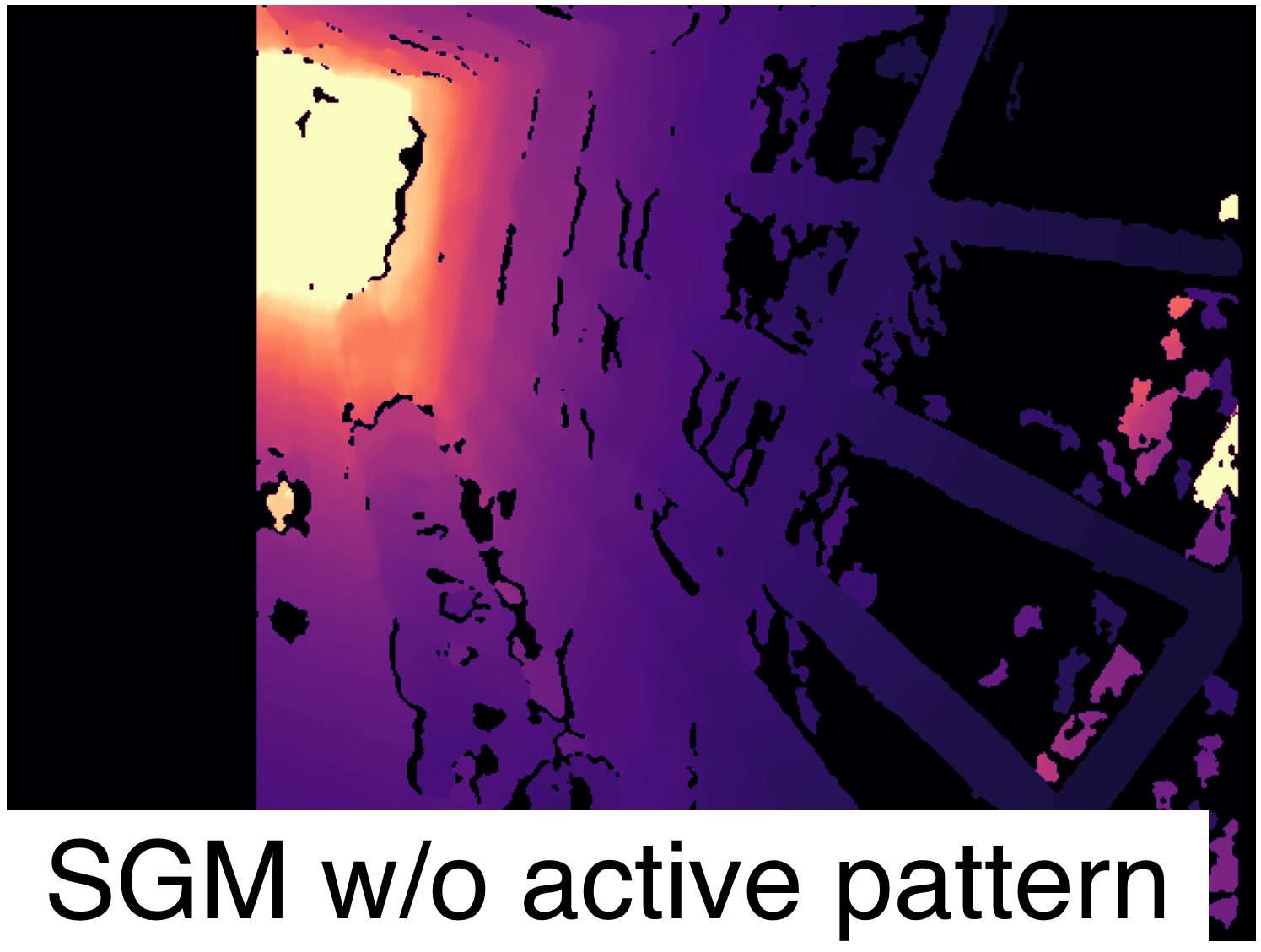}
\end{subfigure}
\begin{subfigure}{.45\linewidth}
  \centering
  \includegraphics[width=\linewidth]{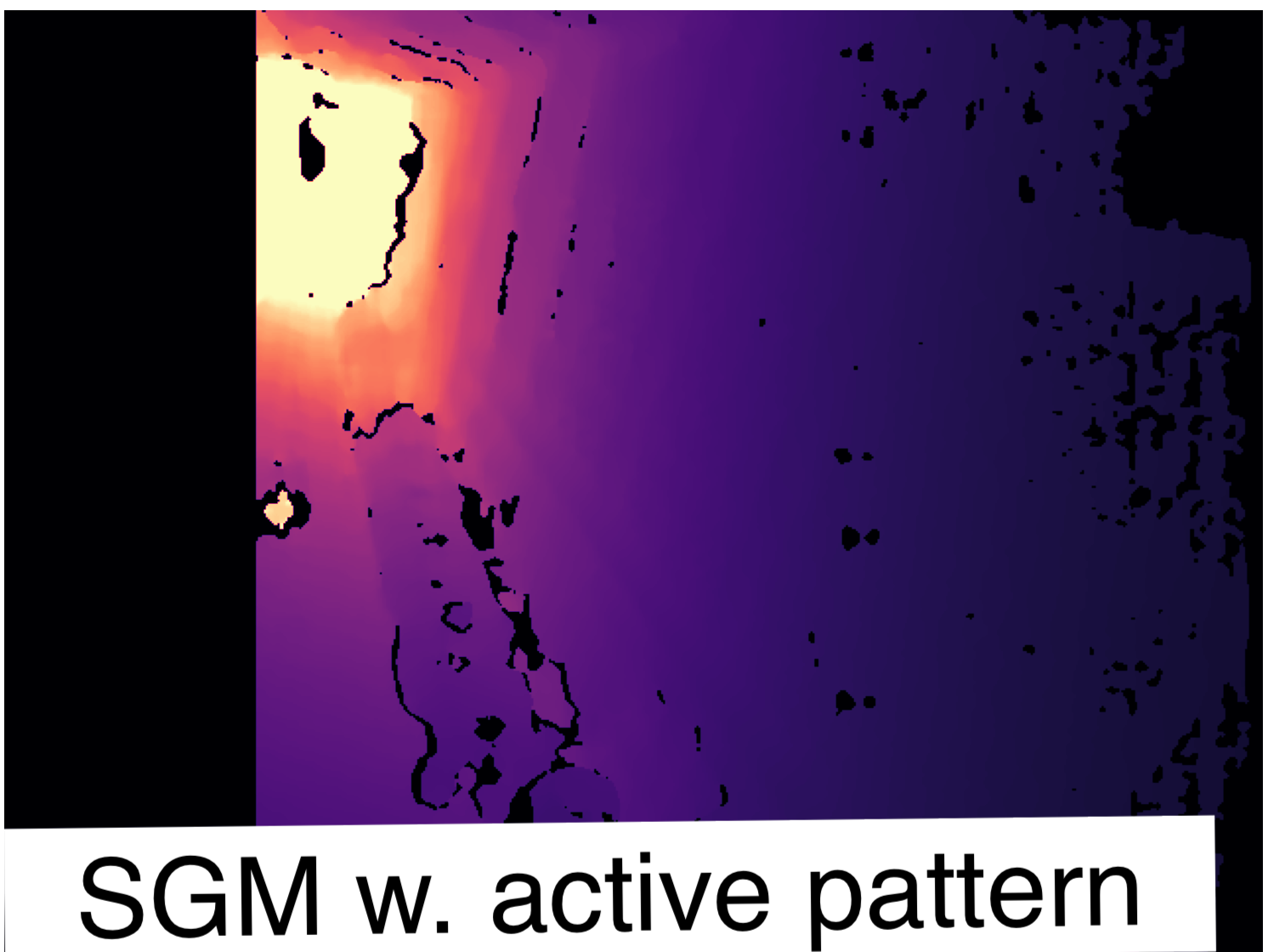}
\end{subfigure}%
\caption{Shows the left stereo image overlaid with the projected texture, the ground truth depth map, and the depth map from SGM~\cite{Hirschmuller08pami} without and with active pattern.}

\label{fig:tartanair_dataset}
\vspace{-0.6cm}
\end{figure}

We use the SGM implementation from OpenCV to create initial depth maps similar to the ones produced by the D435i camera. Fig.~\ref{fig:tartanair_dataset} shows the active texture projected into the left frame, the initial depth map without and with the active pattern and the ground truth depth. 
As this paper is focused on indoor depth completion, we select the four indoor environments 
and one environment that is half indoor and half outdoor.
We two environments for testing, and the remaining three environments for training. We uniformly sample $4979$ frames from the hospital Japanese alley environment, while keeping all $33485$ frames in the training set.

\noindent \textbf{D435i active sequences} is a new dataset for active stereo prediction and completion. The dataset is recorded with multiple D435i sensors across several office and apartment environments. We split the dataset into $15446$ training images, $2762$ validation and $4568$ testing images with no sequence environment overlapping in the training and testing sequences. The D435i records data in an interleaved mode, such that the active pattern is projected onto every second frame. We use the frames without the projected pattern to estimate pseudo ground truth depth maps using the general purpose SfM library COLMAP~\cite{Schoenberger16cvpr,Schoenberger16eccv} that was successfully used to create depth prediction datasets, \eg~\cite{Li18cvpr}. The estimated pseudo ground truth depth maps are semi-dense, but accurate. The depth maps are projected onto the frames with the active pattern, such that we can evaluate the performance on this type of frames.

On average we tracked $240$ 3D landmarks per IR frame on Active TartanAir sequences and $270$ on D435i sequences.

\subsection{Experimental Setup}
\vspace{-0.1cm}

\noindent \textbf{Implementation details.} 
We use an encoder-decoder architecture with skip connections similar to \cite{Wang20arxiv} (using RefineNet \cite{LinLMSR20}), where the encoder is either a ResNet18 (R18) or ResNet50 (R50). The final layer is a sigmoid that predicts normalised disparity $\hat{D}$, which we rescale to absolute depth $1 / (D_{min}  +  (D_{max} -D_{min})\hat{D})$ where $D_{min}$ and $D_{max}$ are hyper parameters for the minimum and maximum disparities. For both datasets we set $D_{min} = 1/20$ and $D_{max} = 1/0.3$. We use the Adam optimizer with $\beta_1 = 0.1$, $\beta_2 = 0.999$, learning rate (lr) $0.00001$, and mini-batch size $12$. We use a lr scheduler that reduces the lr every $15^{th}$ epoch with a factor $10$. The models are trained on a single Nvidia GeForce RTX 3090. Training typically converges after $30$ epochs.

\noindent\textbf{Augmentations.} We perform the following training augmentations, with 50\% chance: random brightness (ranges from 0.8 to 1.2), horizontal flips and random rotation (ranges from -5 to 5 degrees). Augmentations are only applied to the images which are fed to the networks, not to those used to compute losses, where only valid pixels are used. Also, we use Dropblock \cite{GhiasiLL18}, after the first and second ResNet blocks with a drop probability of $0.1$ (disabled during the first training epoch).

\noindent \textbf{Benchmarks.} We benchmark our approach against competing methods for classical stereo~\cite{Geiger10accv}, edge-aware smoothing~\cite{Barron16CVPR}, supervised depth completion~\cite{Senushkin20arxiv}, self-supervised active depth-prediction~\cite{Zhang18eccv} and self-supervised depth-completion~\cite{Wong20icra}. For~\cite{Senushkin20arxiv,Wong20icra} we use the publicly available code with the same implementation details (learning-rate, $\#$ of epochs, optimiser parameters) as proposed in their work. 

As~\cite{Zhang18eccv} do not provide their source code, we implement this closely following the authors instructions. We noticed that the convergence of \ASN~suffered when using only the weighted-LCN loss and instead use a combination of photometric + weighted-LCN loss. According to direct conversation with the authors this helps better generalisation. 

\noindent \textbf{Evaluation Metrics.} Following existing depth completion methods on indoor scenes~\cite{Ma17icra,Senushkin20arxiv,Sinha20eccv} we evaluate each method using the following metrics:
Root mean squared error (RMSE), Mean absolute relative error (Rel.) and $\delta_i^{'s}$ representing the percentage of predicted pixels with relative error below threshold $i$ ($i \in\{1.25,1.25^2,1.25^3\}$). 
We provide the percentage of valid pixels ($\%val$) predicted by each method. 

We split our results into regions with and without initial estimates together to better understand how these methods perform on regions where there was no depth information.

\begin{figure*}[htp!]
    \vspace{0.2cm}
     \centering
     \begin{subfigure}[b]{0.13\textwidth}
         \centering
         \includegraphics[width=\textwidth]{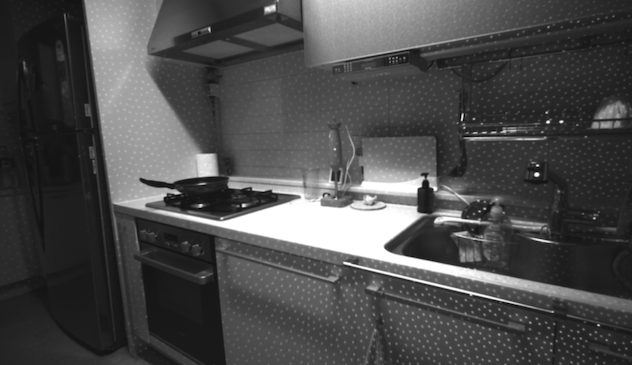}
     \end{subfigure}
     \begin{subfigure}[b]{0.13\textwidth}
         \centering
         \includegraphics[width=\textwidth]{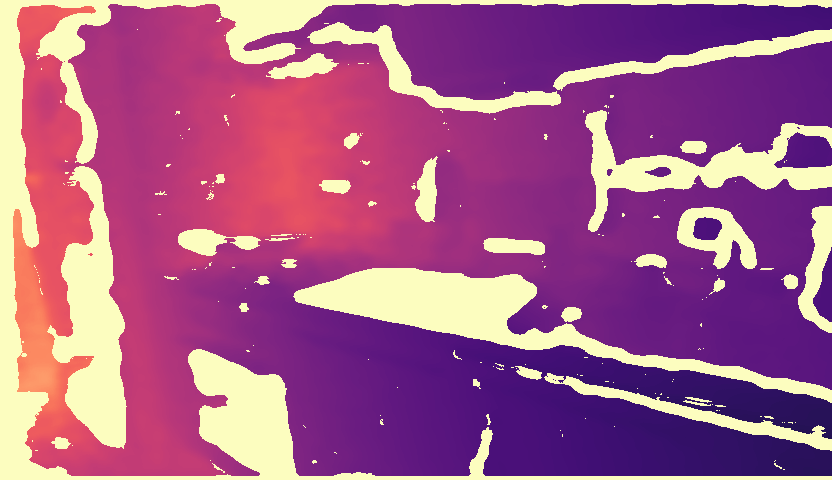}
     \end{subfigure}
     \begin{subfigure}[b]{0.13\textwidth}
         \centering
         \includegraphics[width=\textwidth]{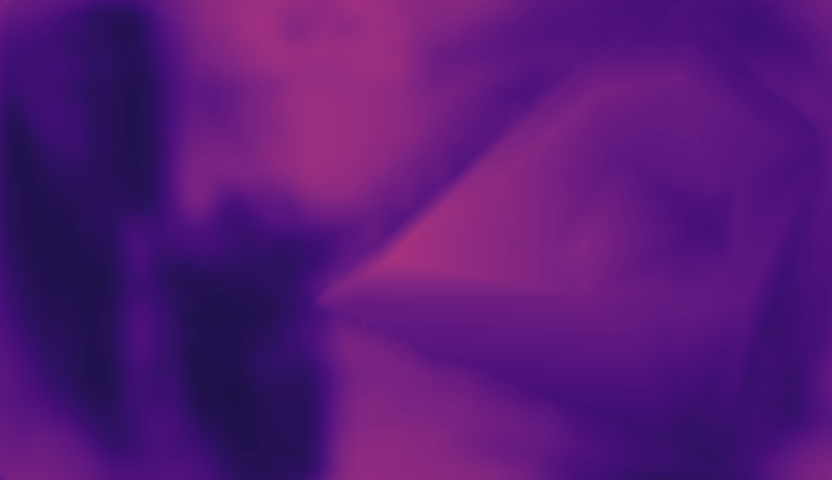}
     \end{subfigure}
     \begin{subfigure}[b]{0.13\textwidth}
         \centering
         \includegraphics[width=\textwidth]{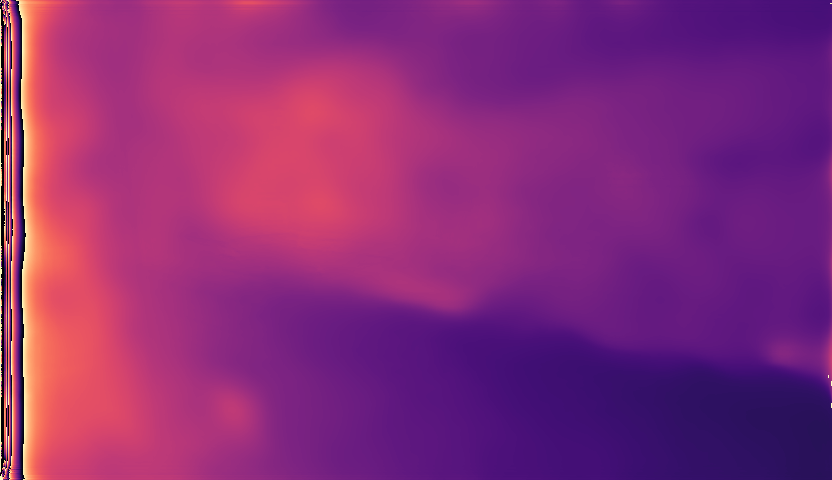}
     \end{subfigure}
     \begin{subfigure}[b]{0.13\textwidth}
         \centering
         \includegraphics[width=\textwidth]{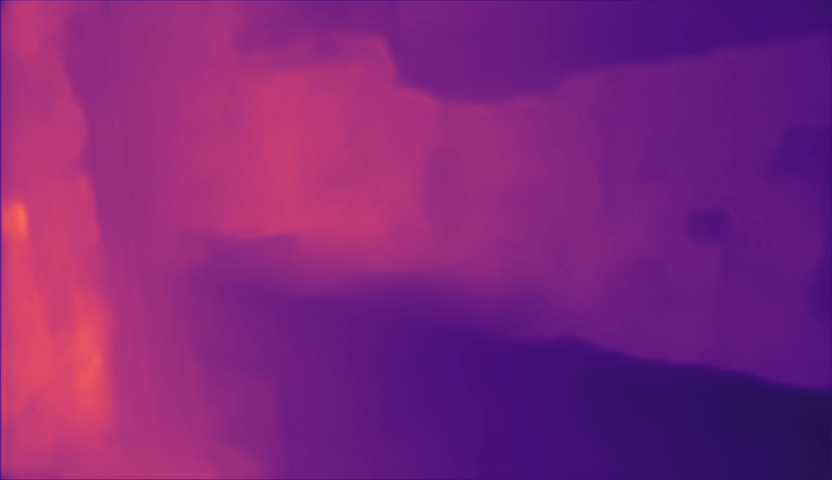}
     \end{subfigure}
     \begin{subfigure}[b]{0.13\linewidth}
         \centering
         \includegraphics[width=\textwidth]{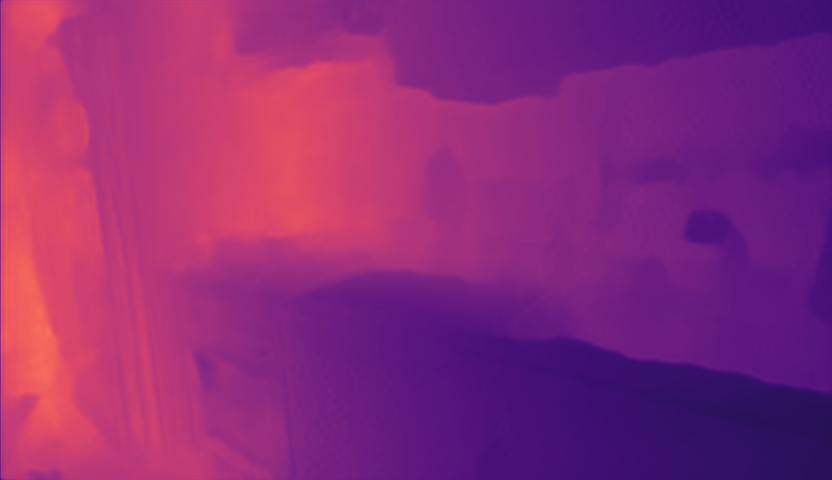}
     \end{subfigure}
     \begin{subfigure}[b]{0.13\textwidth}
         \centering
         \includegraphics[width=\textwidth]{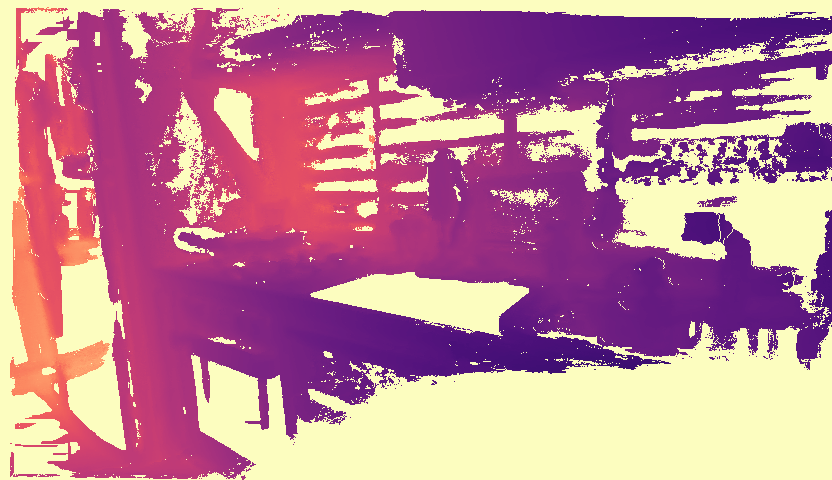}
     \end{subfigure}\\ \vspace{0.1cm}
     \begin{subfigure}[b]{0.13\textwidth}
         \centering
         \includegraphics[width=\textwidth]{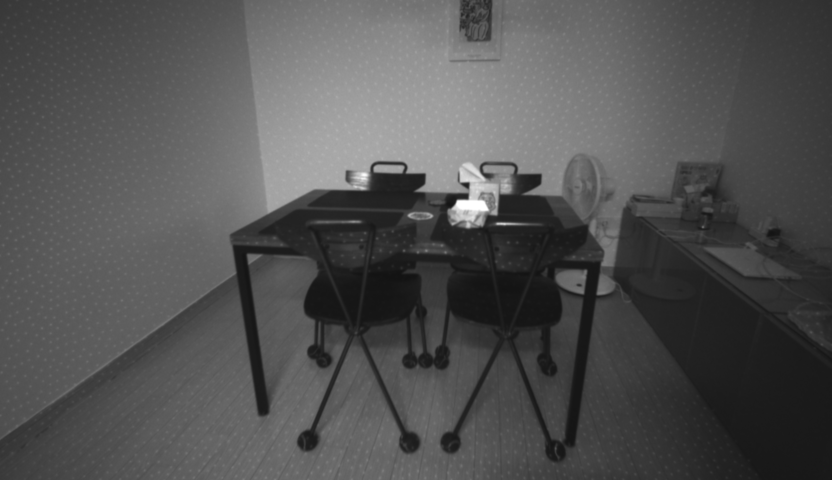}
     \end{subfigure}
     \begin{subfigure}[b]{0.13\textwidth}
         \centering
         \includegraphics[width=\textwidth]{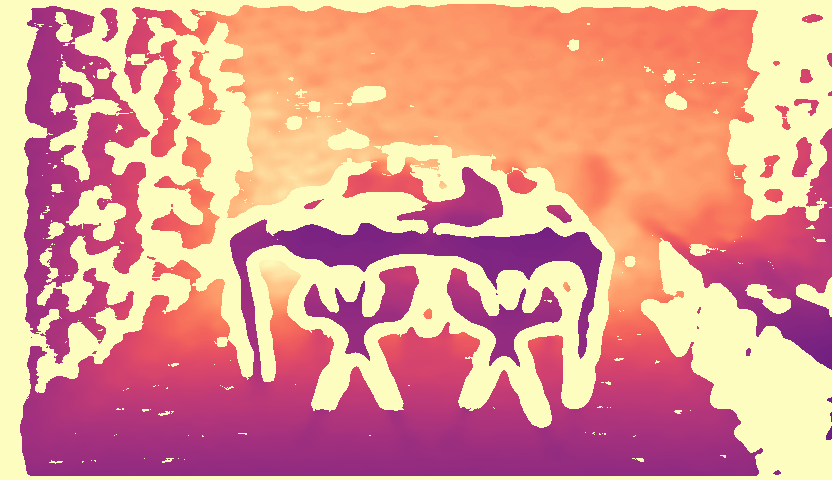}
     \end{subfigure}
     \begin{subfigure}[b]{0.13\textwidth}
         \centering
         \includegraphics[width=\textwidth]{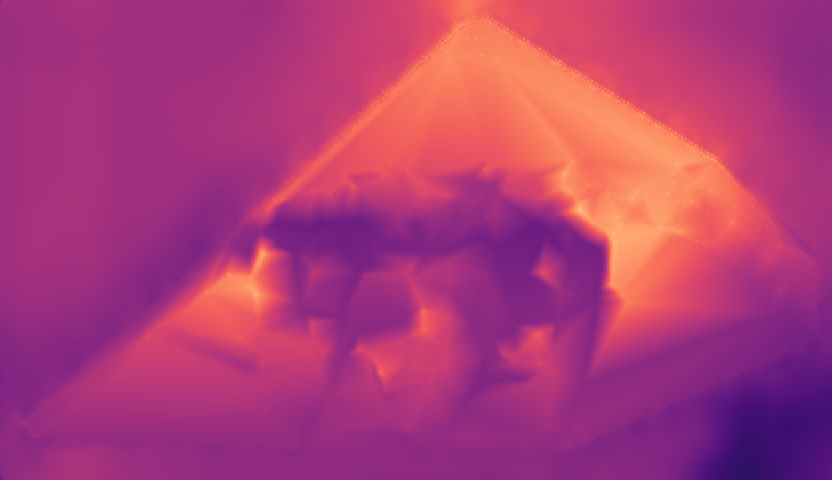}
     \end{subfigure}
     \begin{subfigure}[b]{0.13\textwidth}
         \centering
         \includegraphics[width=\textwidth]{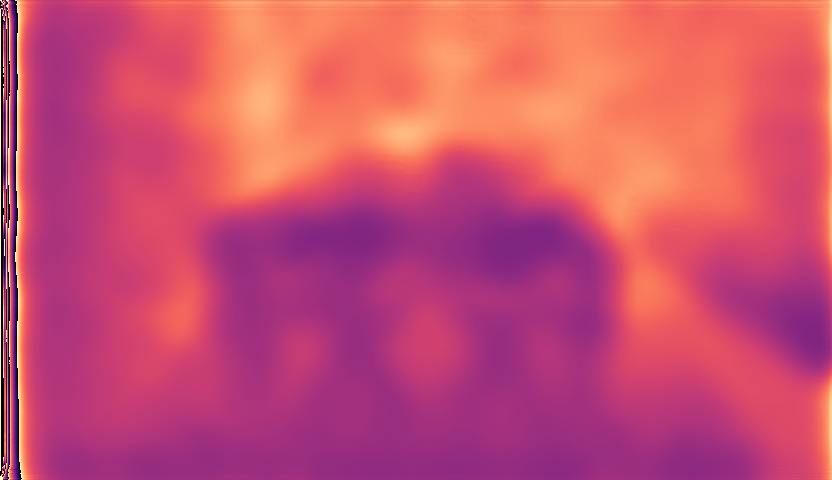}
     \end{subfigure}
     \begin{subfigure}[b]{0.13\textwidth}
         \centering
         \includegraphics[width=\textwidth]{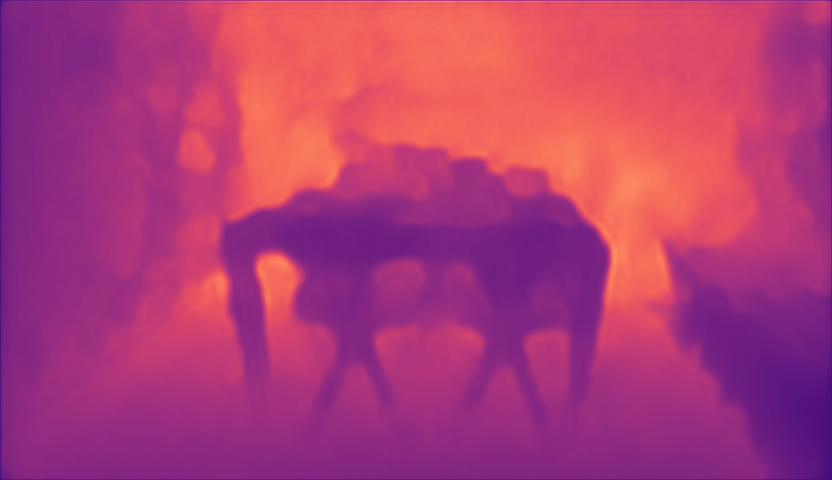}
     \end{subfigure}
     \begin{subfigure}[b]{0.13\linewidth}
         \centering
         \includegraphics[width=\textwidth]{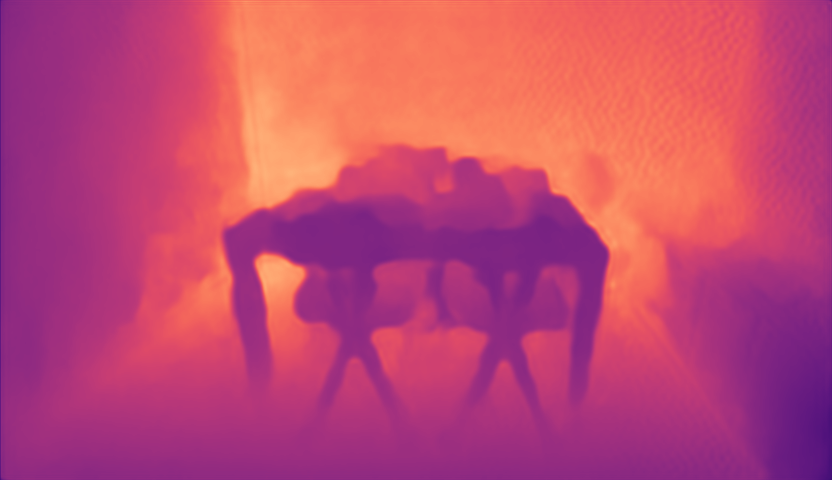}
     \end{subfigure}
     \begin{subfigure}[b]{0.13\textwidth}
         \centering
         \includegraphics[width=\textwidth]{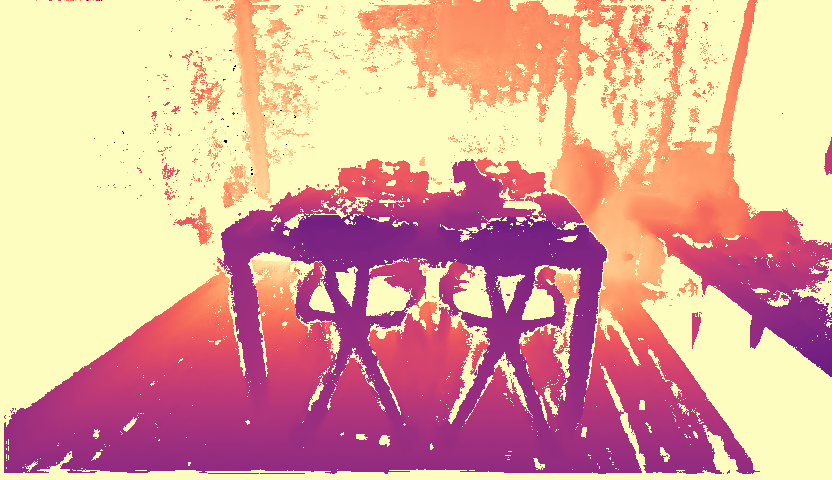}
     \end{subfigure}
     \caption{\textit{From left to right: Input IR0, input depth, VOICED, \ASN, \ACDC-R18, \ACDC-R50 and Ground Truth.} Qualitative comparison shows that our model produces better depth than other self-supervised methods.}
     \vspace{-0.3cm}
     \label{fig:results_d435i}
\end{figure*}

\begin{table*}[htp!]
 \vspace{0.2cm}
  \begin{center}
    \resizebox{\textwidth}{!}{\begin{tabular}{|c|c||c|c|c|c|c|c||c|c||c|c||c|}
      \hline
      \multirow{2}{*}{\textbf{Method}} & 
      \multirow{2}{*}{\textbf{Sup.}} & 
      \multicolumn{6}{c||}{\textbf{Result on whole image}} & 
       \multicolumn{2}{c||}{\textbf{With initial depth}} & 
       \multicolumn{2}{c||}{\textbf{W/O initial depth}} & 
       \textbf{time} $\big\downarrow$\\
       \cline{3-12} 
       &
       & Rel. $\downarrow$& RMSE $\downarrow$& $\delta_1$ $\uparrow$& $\delta_2$ $\uparrow$& $\delta_3$ $\uparrow$& $\%val$ $\uparrow$
       & Rel. $\downarrow$& RMSE $\downarrow$
       & Rel. $\downarrow$& RMSE $\downarrow$
       & (ms) \\
       \hline
       \hline
       
       SGM ~\cite{Hirschmuller08pami} & \xmark
       & 0.176 & 1.549 & 0.773 & 0.789 & 0.808 & 77.7
       & \textbf{0.023} & 0.505 
       & - & - 
       & - \\
       
       ELAS (Robotics)~\cite{Geiger10accv} & \xmark
       & 0.120 & 1.483 & 0.861 & 0.875 & 0.885 & 77.7
       & 0.070 & 1.086 
       & 0.337 & 2.759
       & - \\
       
      Bilateral Solver~\cite{Barron16CVPR} & \xmark
      & 0.190 & \textbf{0.568} & 0.905 & 0.952 & 0.969 & \textbf{100} 
      & 0.062 & \textbf{0.393} 
      & 0.326 & \textbf{0.880}
      & - \\

       \ASN~\cite{Zhang18eccv} & \xmark
       & 0.158 & 1.377 & 0.810 & 0.853 & 0.879 & 86.4
       & 0.110 & 1.182 
       & 0.296 & 2.093 
       & \update{31} \\
       

       \ACDC-R18 (\textbf{ours}) & \xmark
       & 0.130 & 1.049 & 0.875 & 0.954 & 0.977 & \textbf{100}
       & 0.096 & 0.875 
       & 0.215 & 1.616 
       & \update{\textbf{29}} \\

      \update{\ACDC-R50 (\textbf{ours})} & \xmark
    & \update{\textbf{0.087}} & \update{0.805} & \update{\textbf{0.932}} &   \update{\textbf{0.964}} &   \update{\textbf{0.979}} & \update{\textbf{100}}
    & \update{0.037} & \update{0.558} 
    & \update{\textbf{0.174}} & \update{1.416}
    & \update{135} \\
      \hline
      \hline
       
       DMNet~\cite{Senushkin20arxiv} & \cmark
       & 0.110 & 1.217 & 0.846 & 0.933 & 0.968 & 100
       & 0.103 & 1.170 
       & 0.144 & 1.532 
       & \update{38} \\
       
       \hline
   \end{tabular}}
  \end{center}
  \caption{We compare \ACDC~with state of the art methods on both regions w./w.o initial depth estimates in \textbf{Active TartanAir} sequences. \textbf{Note} DMNet is the only supervised method (\textbf{Sup.}) in this table and hence shown separately. 
  We observe that despite being a self-supervised method, our work closes the gap between supervised and self-supervised methods while been computationally efficient. Inference times are for an image resolution of $832\times480$ on RTX 3090. \textbf{Bold} values are best results for the self-supervised methods. $\downarrow$ shows metrics that are better with lower values and $\uparrow$ is vice-versa.}\label{tab:tartanair}
  \vspace{-0.3cm}
\end{table*}

\subsection{Results}
\begin{figure}
\vspace{0.2cm}
     \centering
     \begin{subfigure}{0.18\linewidth}
         \centering
         \includegraphics[width=\textwidth]{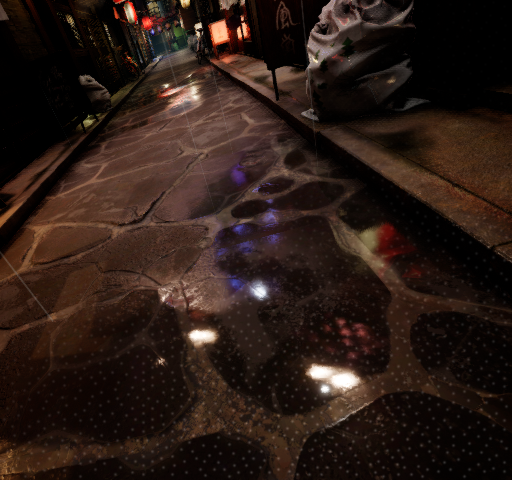}
     \end{subfigure}
     \begin{subfigure}{0.18\linewidth}
         \centering
         \includegraphics[width=\textwidth]{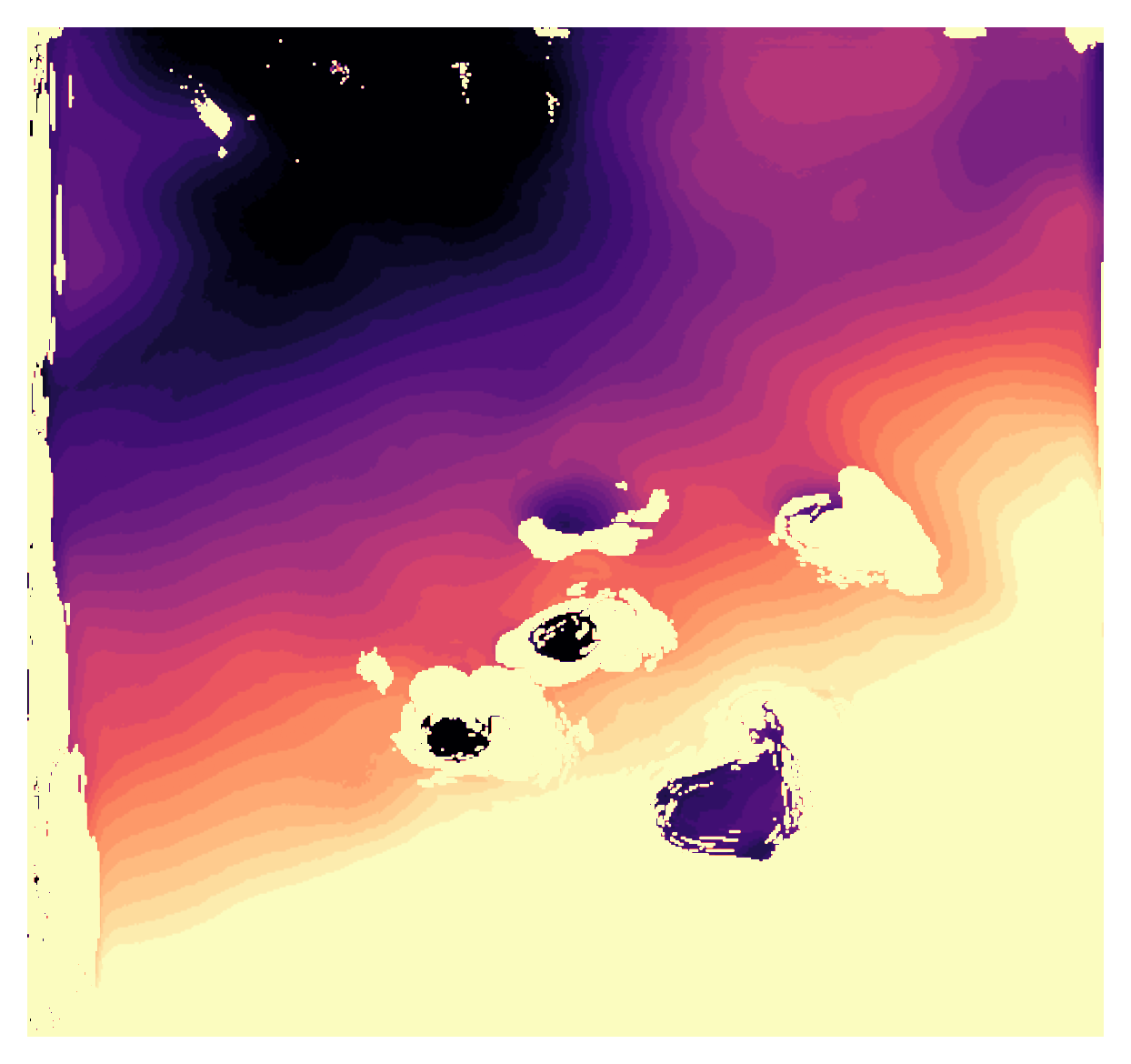}
     \end{subfigure}
     \begin{subfigure}{0.18\linewidth}
         \centering
         \includegraphics[width=\textwidth]{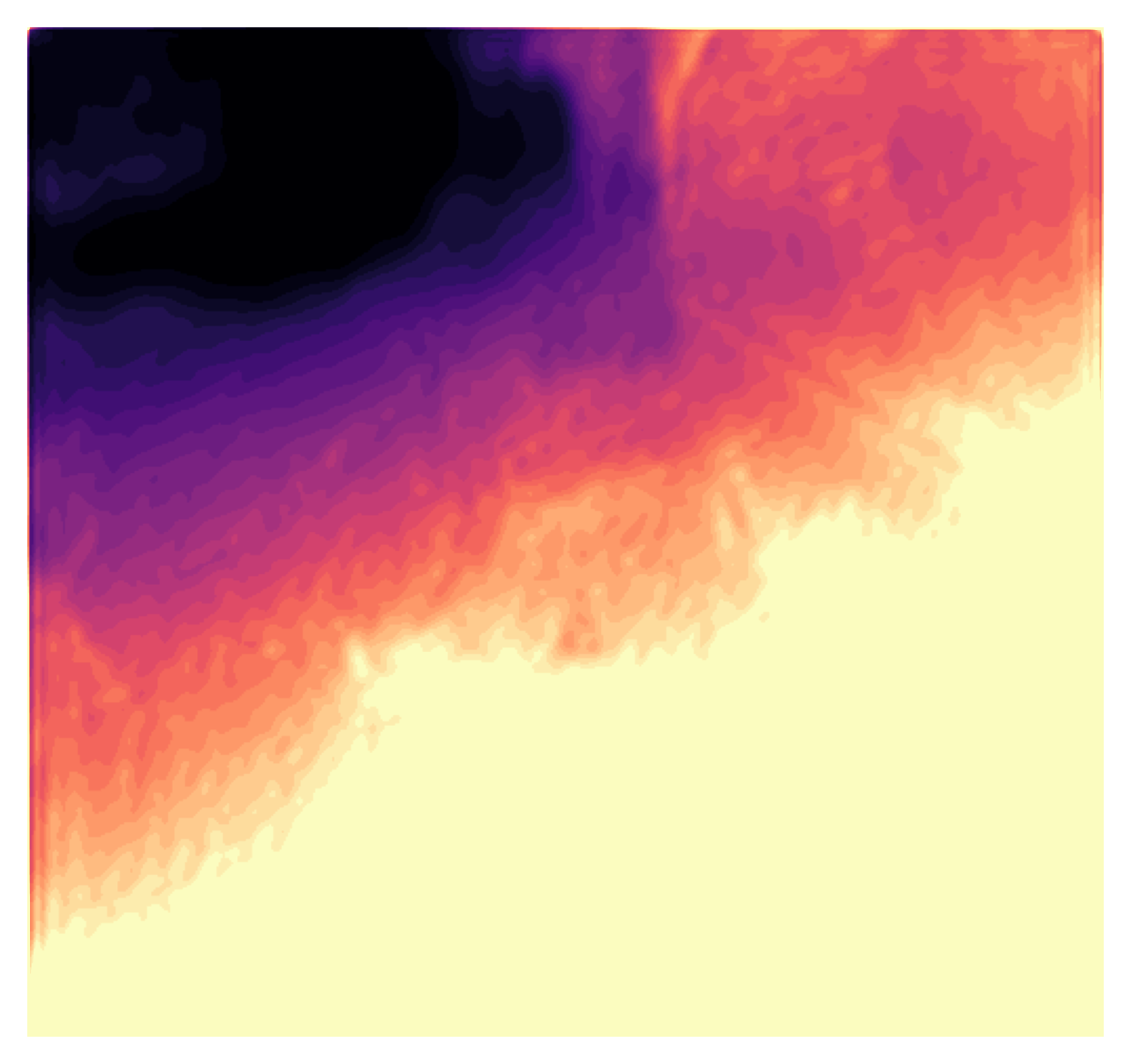}
     \end{subfigure}
     \begin{subfigure}{0.18\linewidth}
         \centering
         \includegraphics[width=\textwidth]{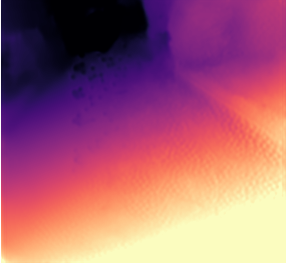}
     \end{subfigure}
     \begin{subfigure}{0.18\linewidth}
         \centering
         \includegraphics[width=\textwidth]{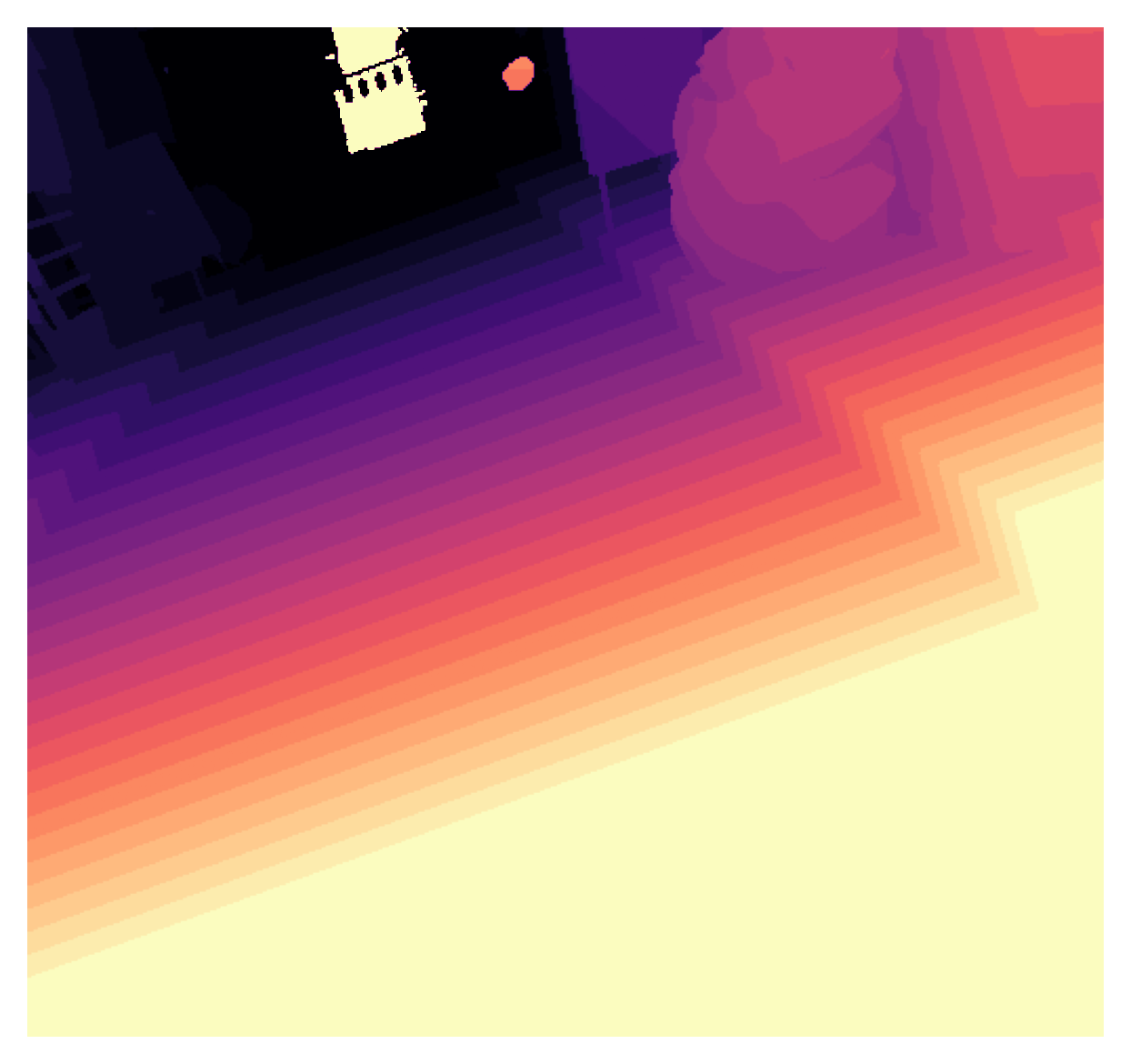}
     \end{subfigure}

    \begin{subfigure}[b]{0.18\linewidth}
         \centering
         \includegraphics[width=\textwidth]{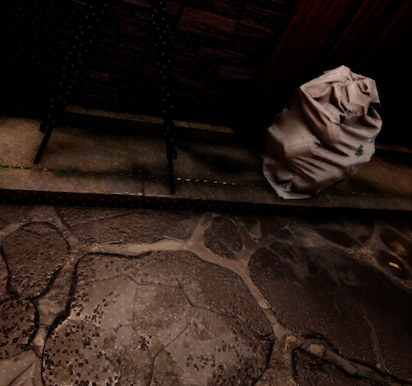}
     \end{subfigure}
     \begin{subfigure}[b]{0.18\linewidth}
         \centering
         \includegraphics[width=\textwidth]{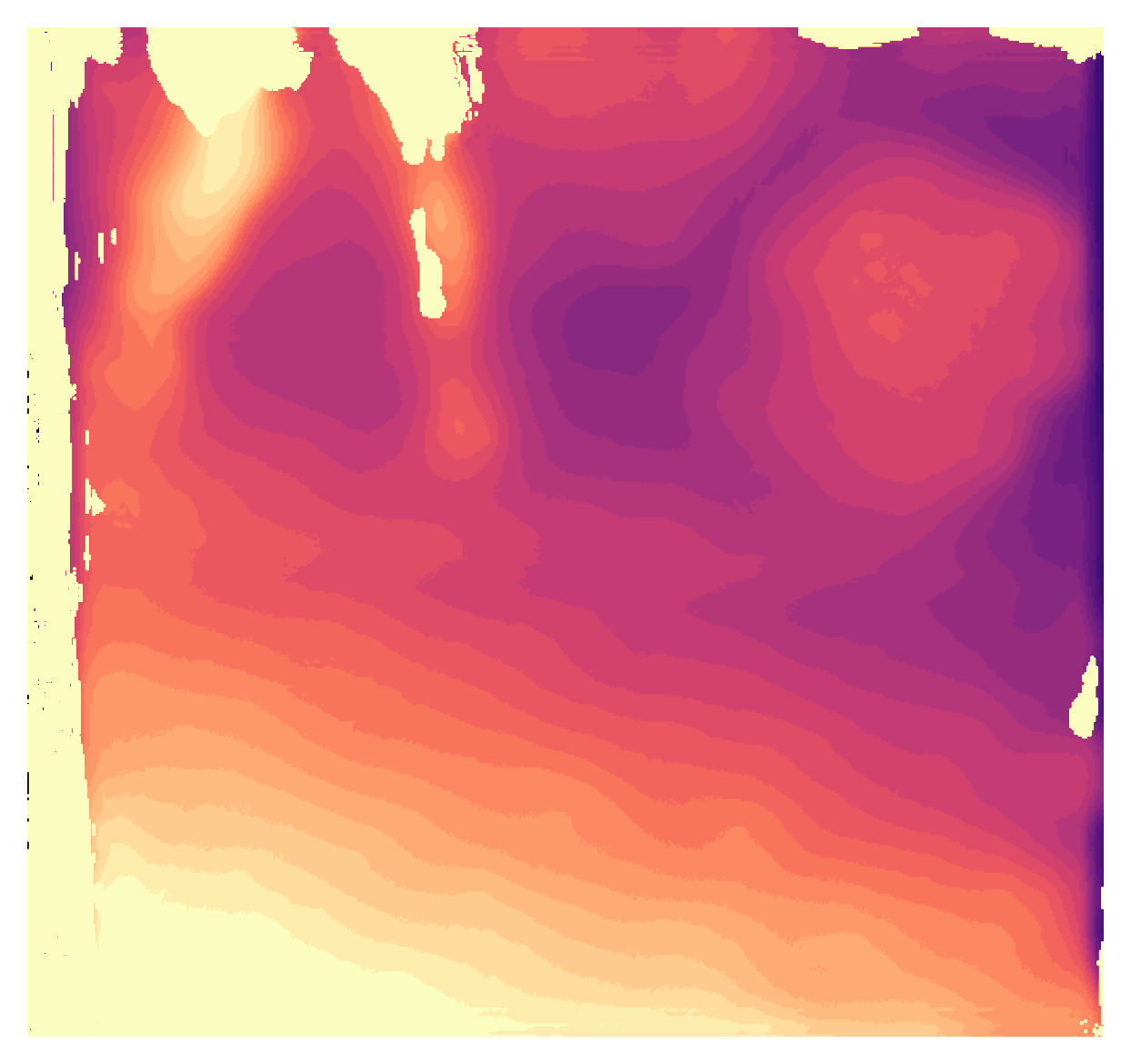}
     \end{subfigure}
     \begin{subfigure}[b]{0.18\linewidth}
         \centering
         \includegraphics[width=\textwidth]{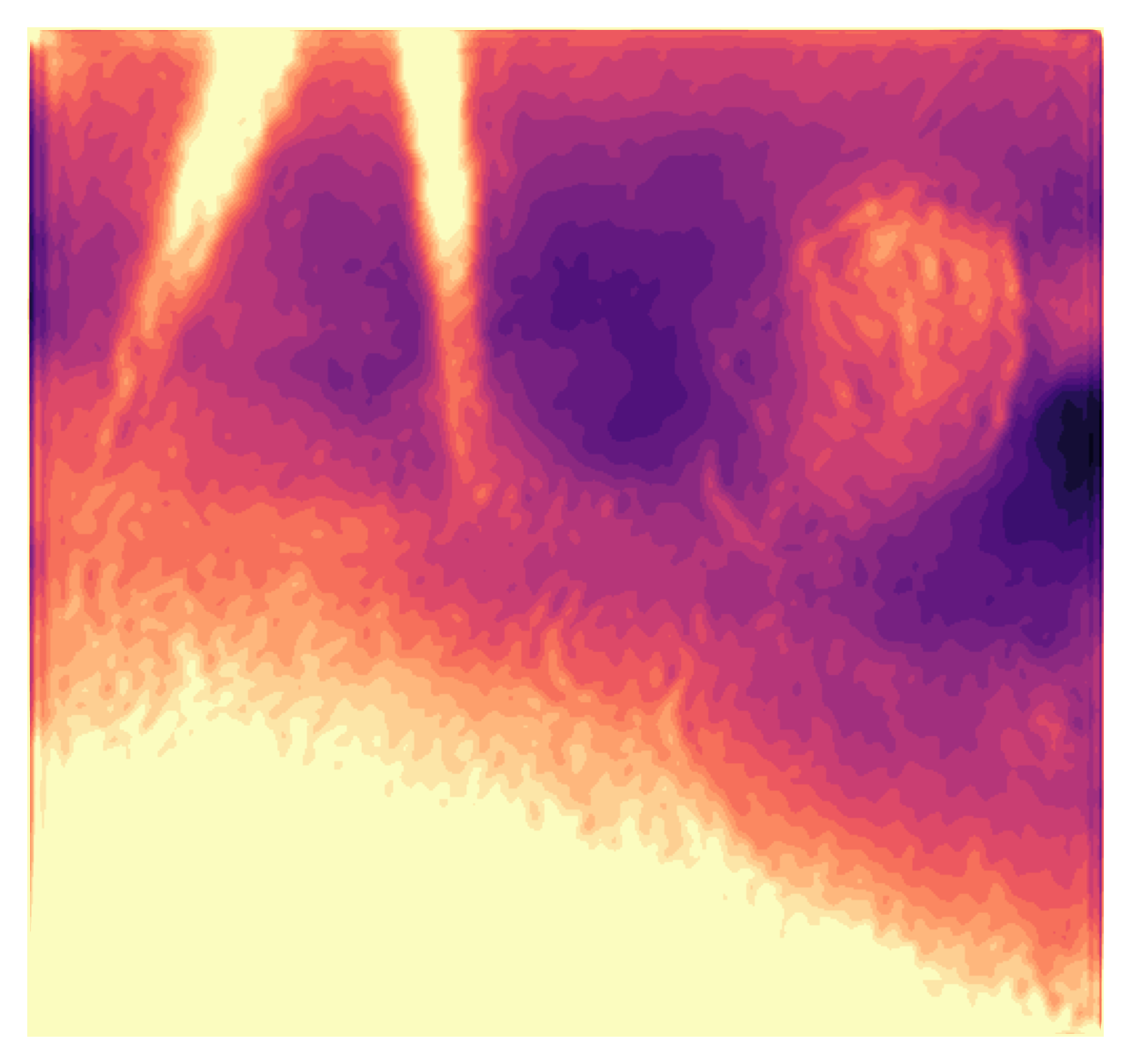}
     \end{subfigure}
     \begin{subfigure}[b]{0.18\linewidth}
         \centering
         \includegraphics[width=\textwidth]{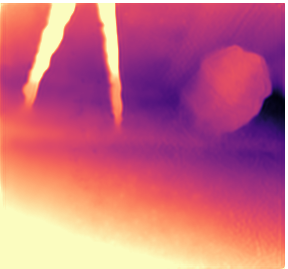}
     \end{subfigure}
     \begin{subfigure}[b]{0.18\linewidth}
         \centering
         \includegraphics[width=\textwidth]{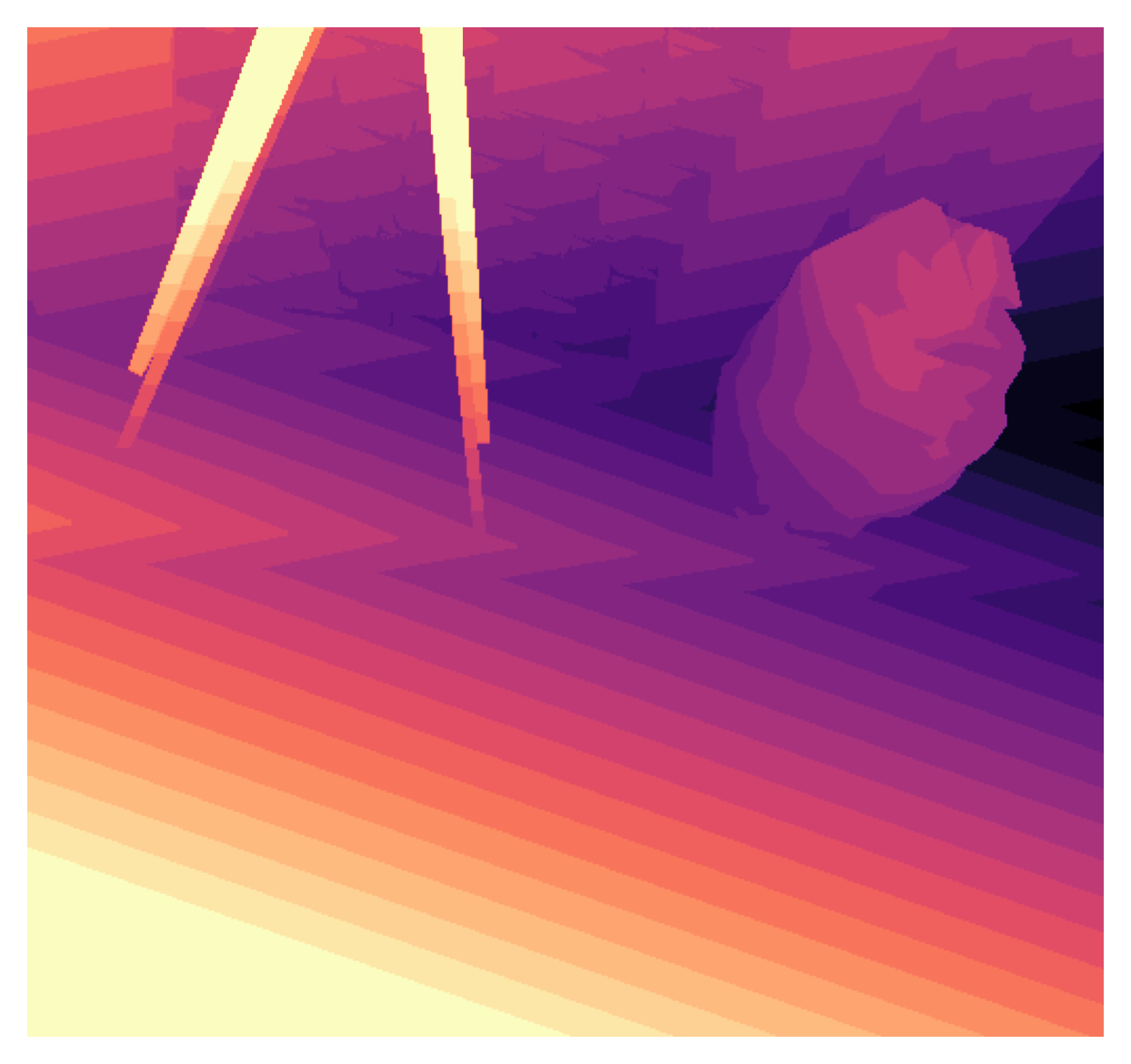}
     \end{subfigure}

     \begin{subfigure}[b]{0.18\linewidth}
         \centering
         \includegraphics[width=\textwidth]{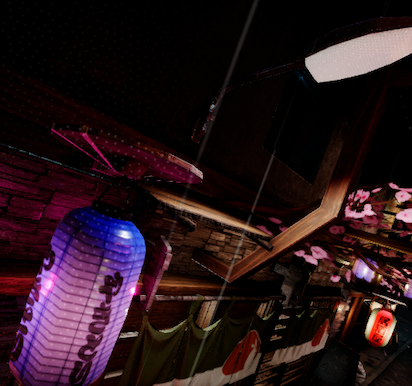}
     \end{subfigure}
     \begin{subfigure}[b]{0.18\linewidth}
         \centering
         \includegraphics[width=\textwidth]{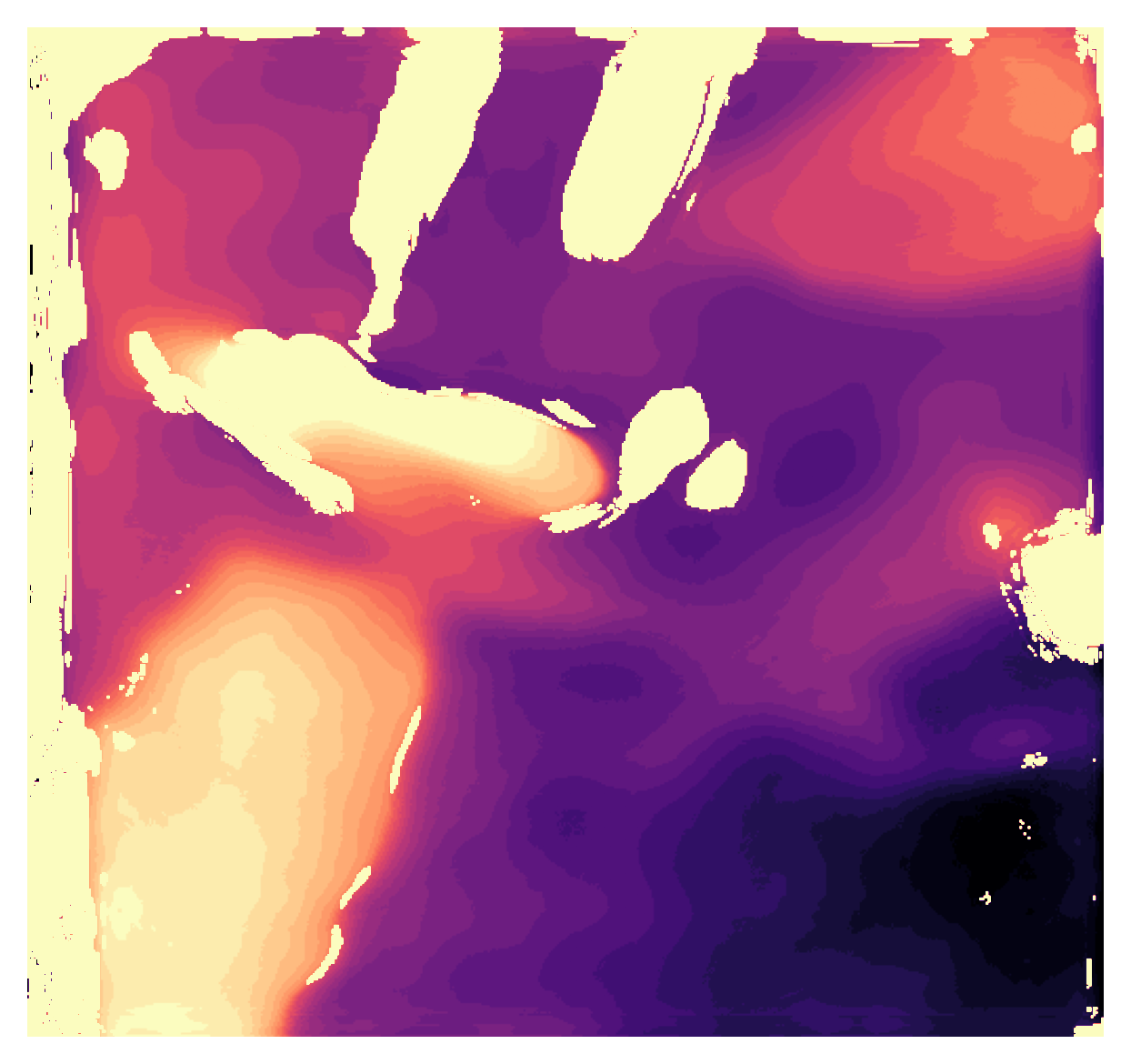}
     \end{subfigure}
     \begin{subfigure}[b]{0.18\linewidth}
         \centering
         \includegraphics[width=\textwidth]{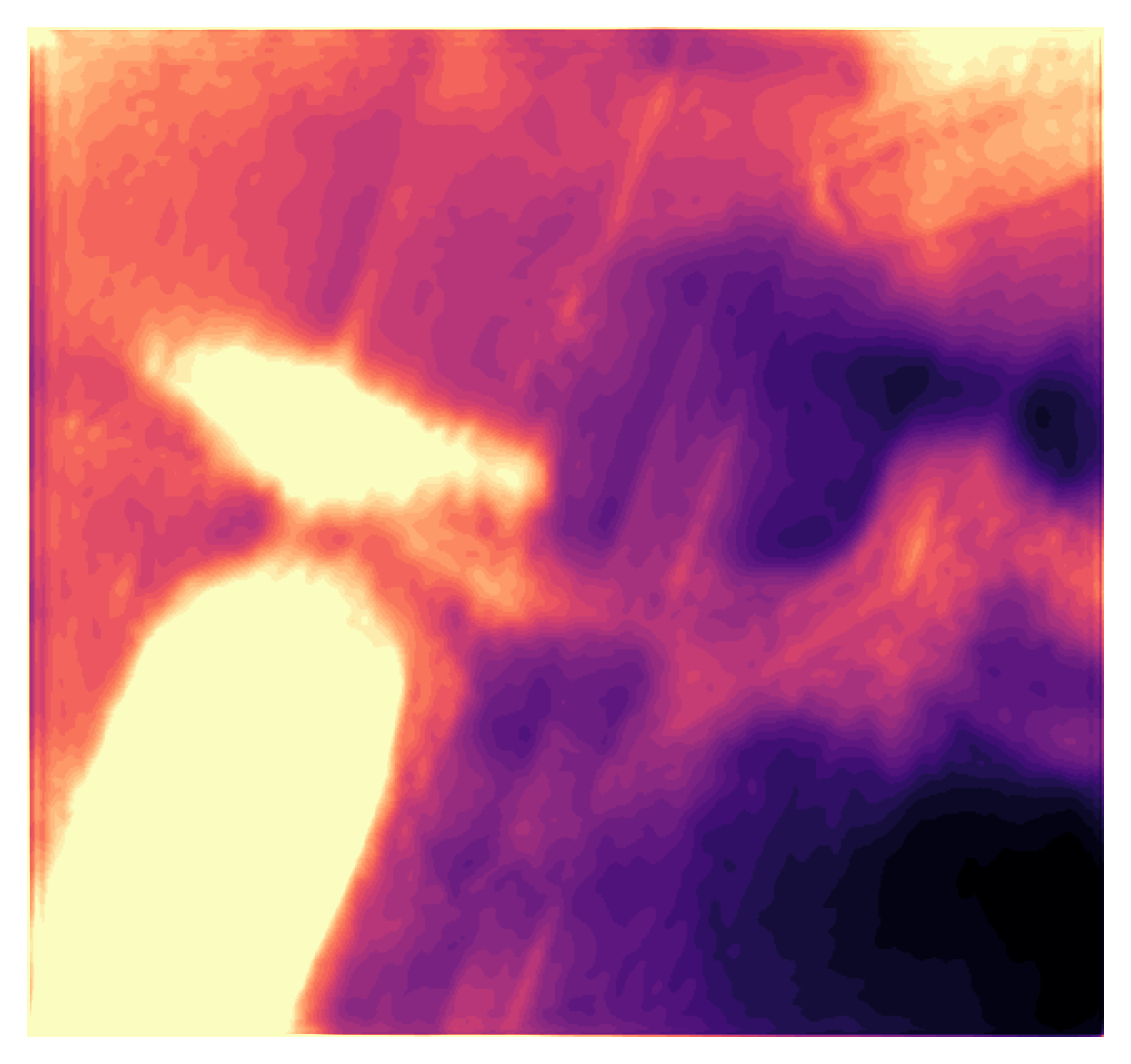}
     \end{subfigure}
     \begin{subfigure}[b]{0.18\linewidth}
         \centering
         \includegraphics[width=\textwidth]{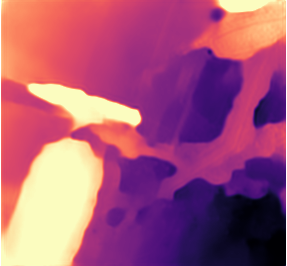}
     \end{subfigure}
     \begin{subfigure}[b]{0.18\linewidth}
         \centering
         \includegraphics[width=\textwidth]{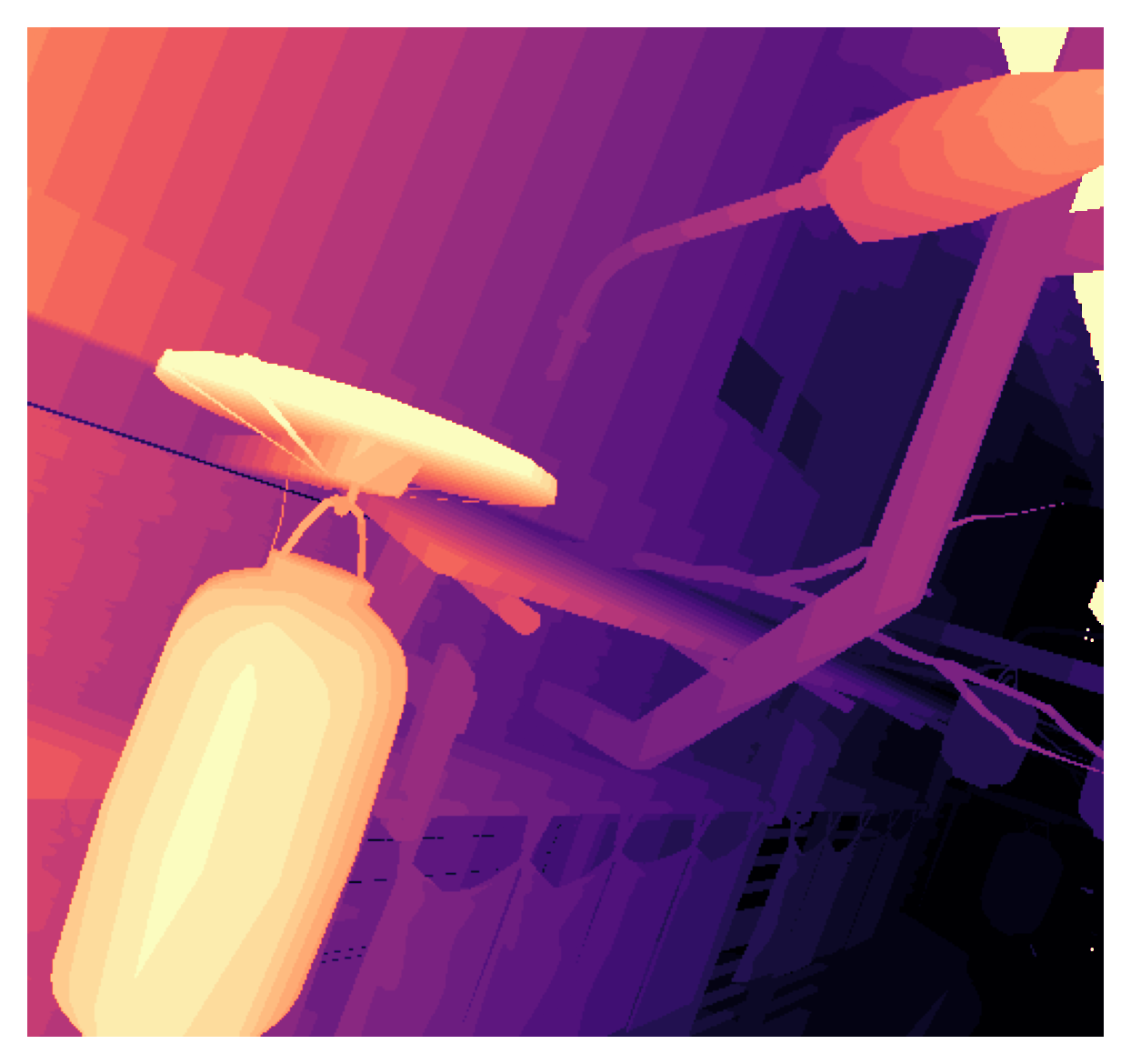}
     \end{subfigure}
     
     \caption{\textit{From left to right: Input RGB, \ASN, \ACDC-R18, \update{\ACDC-R50} and Ground Truth.} Qualitative comparison shows that our model produces accurate depth for challenging scenarios as specularities ($1^{st}$ row), thin structures ($2^{nd}$ row), rain drops ($3^{th}$ row).}
    \vspace{-0.5cm}
     \label{fig:results_tartanair}
\end{figure}

\vspace{-0.1cm}
\noindent\textbf{Active TartanAir.} In Table~\ref{tab:tartanair} we compare our work against several baseline methods on the Active TartanAir dataset. While SGM does well where it predicts depth, it fails to produce estimates in texture-less regions without active pattern. Moreover, due to the wide baseline in this dataset, large occluded regions are present in the stereo images leading to low completeness score for both \ASN~and SGM. \ACDC~outperforms competing methods, particularly in regions without any initial depth prediction. In contrast to SGM, \ACDC exploits the combination of guide images and sparse 3D landmarks to predict more accurate depth estimates. Additionally, our method is computationally efficient and running at over $25$ FPS. Fig.~\ref{fig:results_tartanair} shows qualitatively that our network produces crisper results compared to other methods. 
\begin{table*}[htp!]
 \small
  \begin{center}
    \resizebox{\textwidth}{!}{\begin{tabular}{|c|c|c|c||c|c|c|c|c|c||c|c||c|c|}
      \hline
      \multirow{2}{*}{\textbf{Method}} &
      \multirow{2}{*}{\textbf{S}} & \multirow{2}{*}{\textbf{C}} & \multirow{2}{*}{\textbf{A}} &
      \multicolumn{6}{c||}{\textbf{Result on whole image}} & 
       \multicolumn{2}{c||}{\textbf{With initial depth}} & 
       \multicolumn{2}{c|}{\textbf{W/O initial depth}} \\
       \cline{5-14} 
       & & & 
       & Rel. $\downarrow$& RMSE $\downarrow$ & $\delta_1$ $\uparrow$& $\delta_2$ $\uparrow$& $\delta_3$ $\uparrow$ & $\%val$ $\uparrow$
       & Rel. $\downarrow$& RMSE $\downarrow$
       & Rel. $\downarrow$& RMSE $\downarrow$
       \\
       \hline
       \hline
       
       SGM ~\cite{Hirschmuller08pami} & \cmark &  &
       & 0.236 & 0.957 & 0.719 & 0.736 & 0.748 & 57.8
       & 0.051 & 0.297
       & - & - 
       \\

       ELAS (Robotics)~\cite{Geiger10accv} & \cmark &  &
       & 0.078 & 0.402 & 0.931 & 0.945 & 0.952 & 84.1
       & \textbf{0.045} & 0.227 
       & 0.193 & 0.715 
       \\

       \ASN~\cite{Zhang18eccv} & \cmark &  & \cmark
       & 0.123 & 0.538 & 0.903 & 0.957 & 0.973 & 96.4
       & 0.066 & 0.261 
       & 0.308 & 0.997 
       \\

       \hline
       
      Bilateral Solver~\cite{Barron16CVPR} &  & \cmark & 
      & 0.090 & 0.307 & 0.931 & 0.974 & 0.984 & 97.7 
      & 0.070 & 0.226 
      & 0.160 & 0.479
      \\
      
    
    \update{S2D-R34~\cite{Ma17icra}} &  & \cmark & \cmark 
    & \update{0.383} & \update{1.008} & \update{0.326}  &   \update{0.507}  &   \update{0.667}  &   \update{\textbf{100}}
    & \update{0.360} & \update{0.982}
    & \update{0.407} & \update{1.074} \\
      
    \update{Concat inputs (R50)} &  & \cmark & \cmark
    & \update{0.126} & \update{0.468} & \update{0.860}  &   \update{0.945}  &   \update{0.970}  &   \update{\textbf{100}} 
    & \update{0.090} & \update{0.323} 
    & \update{0.194} & \update{0.701}
    \\
    
      VOICED~\cite{Wong20icra} &  & \cmark & 
      & 0.194 & 0.569 & 0.737 & 0.874 & 0.934 & 98.7 
      & 0.179 & 0.482 
      & 0.239 & 0.761 
      \\
       

      \update{\ACDC-R50 (\textbf{Mean})} &  & \cmark & \cmark
        & \update{0.128} & \update{0.374} & \update{0.911}  &   \update{0.957}  &   \update{0.973}  &   \update{\textbf{100}}
        & \update{0.101} & \update{0.268} 
        & \update{0.164} & \update{0.556}
        \\
       
       \ACDC-R18 (\textbf{ours}) &  & \cmark & \cmark
      & 0.095 & 0.361 & 0.909 & 0.974 & 0.986 & \textbf{100} 
      & 0.075 & 0.253 
      &  0.148 & 0.550 
      \\
      
      \ACDC-R50 (\textbf{ours}) &  & \cmark & \cmark
      & \textbf{0.075} &\textbf{0.290} & \textbf{0.945} & \textbf{0.981} & \textbf{0.988} & \textbf{100} 
      & 0.055 & \textbf{0.184}
      & \textbf{0.117} & \textbf{0.460}
      \\
       
      \hline
   \end{tabular}}
  \end{center}
  \caption{
  With a combination of inputs and complementary losses \ACDC-\{R18,R50\}~outperforms competing methods on the \textbf{D435i} sequences. We benchmark against \textbf{Stereo} (S) and \textbf{Completion} (C) methods w./w.o. the \textbf{Active} pattern (A).}
  \vspace{-0.3cm}
  \label{tab:realsense}
\end{table*}

\noindent\textbf{D435i.} 
Table~\ref{tab:realsense} shows that \ACDC~outperforms the self-supervised baselines (VOICED, \ASN, \update{S2D}). Similar to the results on Active TartanAir datasets, we see that SGM performs well where it predicts depth. However, \ACDC~exploits the sparse landmarks and together with temporal losses learns better priors even on regions without initial depth predictions. \update{To emphasize the need for the proposed $\max$ operator in the channel exchange framework~\cite{Wang20arxiv}, we compare with two baseline version. In the first one we concatenate all the inputs and feed it into a single R50 backbone (Concat inputs), and for the other we use the vanilla mean operator based channel exchange network. We observe that only when we use the $\max$ operator are we able to convincingly out-perform competing methods producing state of the art active depth completion results.} Fig.~\ref{fig:results_d435i}
shows qualitative performance of \ACDC~R18 compared with VOICED~\cite{Wong20icra} and \ASN~\cite{Zhang18eccv} on the D435i sequences. Note that VOICED is not able to recover from the initial scaffolding.

\begin{figure}
    \vspace{0.2cm}
     \centering
     \begin{subfigure}[b]{0.15\textwidth}
         \centering
         \includegraphics[width=\textwidth]{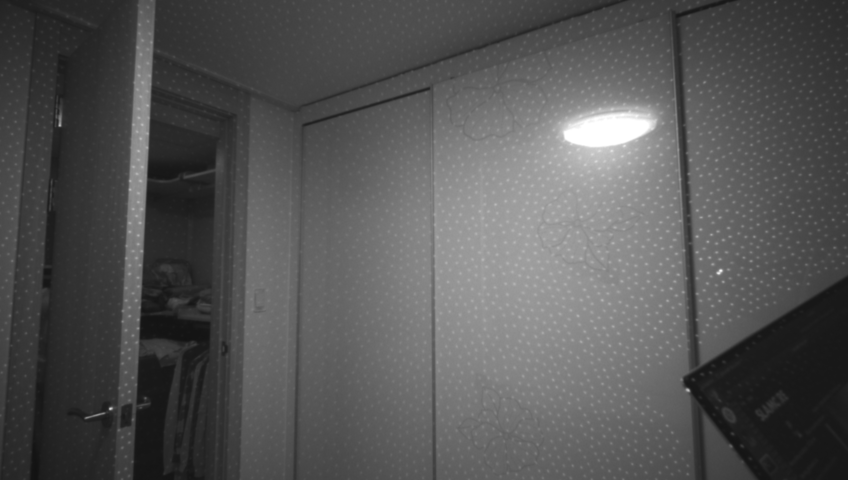}
     \end{subfigure}
     \begin{subfigure}[b]{0.15\textwidth}
         \centering
         \includegraphics[width=\textwidth]{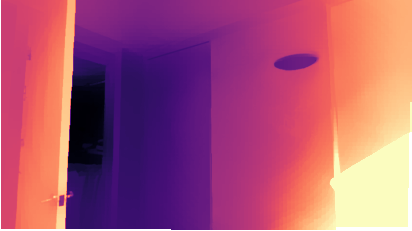}
     \end{subfigure}
     \begin{subfigure}[b]{0.15\textwidth}
         \centering
         \includegraphics[width=\textwidth]{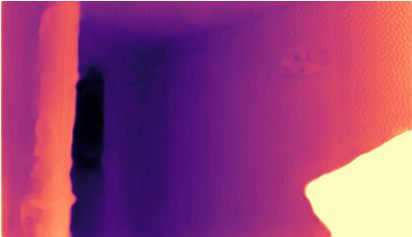}
     \end{subfigure}
     \begin{subfigure}[b]{0.15\textwidth}
         \centering
         \includegraphics[width=\textwidth]{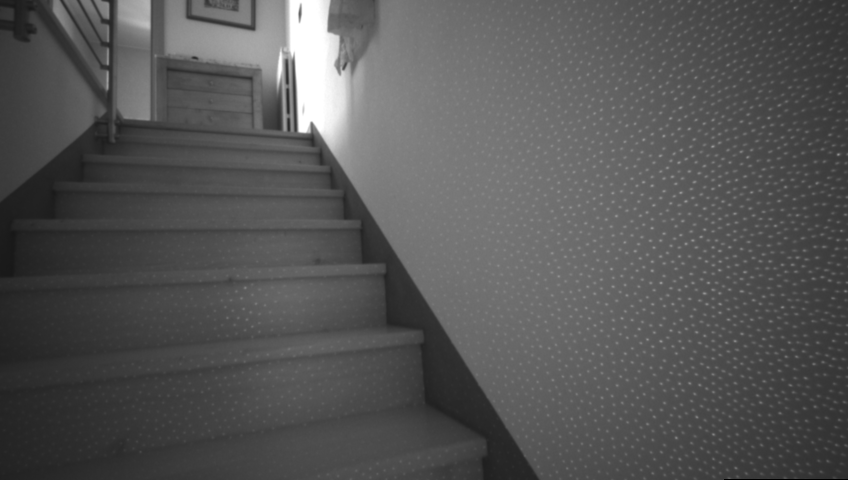}
     \end{subfigure}
     \begin{subfigure}[b]{0.15\textwidth}
         \centering
         \includegraphics[width=\textwidth]{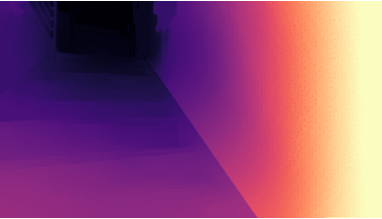}
     \end{subfigure}
     \begin{subfigure}[b]{0.15\textwidth}
         \centering
         \includegraphics[width=\textwidth]{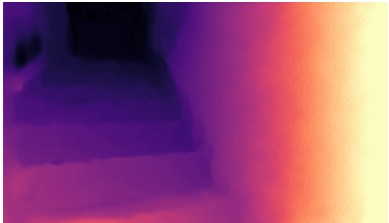}
     \end{subfigure}     
        \caption{\update{\textit{From left to right:} \textit{Input IR, Bilateral solver, \ACDC-R50.} While bilateral solver produces sharp edges, it follows the guide image resulting in false depth discontinuities. \ACDC~extracts scene understanding to interpolate depth.}}
        \label{fig:bilateral_solver}
        \vspace{-0.3cm}
\end{figure}

\begin{figure}[htp!]
     \centering
     \includegraphics[width=\columnwidth]{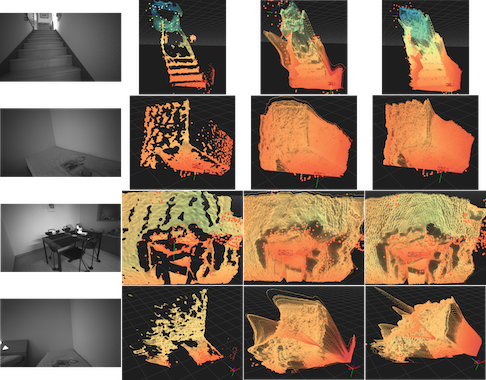}
     \caption{\textit{From left to right:} \textit{Input IR, input depth map, \ACDC-R18, \ACDC-R50.} Point-clouds showing the spatial coherency of our method and its ability to fill missing crucial information of the environment, such as walls.}
     \label{fig:results_realsense}
     \vspace{-0.3cm}
\end{figure}

\update{In Fig.~\ref{fig:bilateral_solver} we observe that while Bilateral solver produces sharper edges compared to \ACDC, it closely follows the guide image resulting in false depth discontinuities from shadows, glare and changes in intensity. In comparison learning based methods like \ACDC~exploit high level scene understanding to avoid them.} We also present the output as point clouds (Fig.~\ref{fig:results_realsense}) to highlight the spatial coherence of our prediction and its potential for downstream tasks, such as robot navigation and scene understanding. Our model is able to complete the initial depth maps with guidance from the IR frame and sparse points even when there are large invalid areas in the input. \ACDC~correctly fills missing depth such as the walls next to the stairs in the first example. 


\noindent\textbf{Ablation Studies.} To investigate the impact of different input signals (sparse, depth and guide-IR) for depth completion we conduct ablation studies on the D435i dataset. Specifically, we train multiple networks with different combinations of input signals and observe their performance. In Table~\ref{tab:ablation_input} we observe that each input contribute to the model performance. The sparse depth especially helps in distant regions (low RMSE).

\begin{table}[htp!]
\vspace{0.2cm}
 \small
  \begin{center}
    \begin{tabular}{|c|c|c|c|c|c|c|c|}
      \hline 
       \emph{IR} & \emph{Dep.} & \emph{Spa.}
       & Rel. $\downarrow$& RMSE $\downarrow$& $\delta_1$ $\uparrow$& $\delta_2$ $\uparrow$& $\delta_3$$\uparrow$
       \\
       \hline
       \hline
       
        & \cmark & \cmark
       & 0.150  &  0.710  & 0.826 & 0.931 & 0.958
       \\
       
       \cmark &  & \cmark
       & 0.165 & 0.574 & 0.749 & 0.925 & 0.968
       \\
       
       \cmark & \cmark & 
       & 0.417 & 0.883 & 0.761 & 0.911 & 0.949
       \\
       
       \cmark & \cmark & \cmark
       & \textbf{0.095} & \textbf{0.361} & \textbf{0.909} & \textbf{0.974} & \textbf{0.986}
       \\
      \hline
   \end{tabular}
  \end{center}
  \caption{Impact of different inputs. \emph{IR} is the reference Guide-IR image, \emph{Dep.} refers to the initial depth and \emph{Spa.} to the projected 3D landmarks.}
  \vspace{-0.3cm}
  \label{tab:ablation_input}
\end{table}

\begin{table}[htp!]
 \small
  \begin{center}
    \begin{tabular}{|c|c|c|c|c|c|c|}
      \hline 
       \emph{Temp.} & \emph{Sparse}
       & Rel. $\downarrow$& RMSE $\downarrow$& $\delta_1$ $\uparrow$& $\delta_2$ $\uparrow$& $\delta_3$$\uparrow$
       \\
       \hline
       \hline
       
       \cmark & 
       & 0.138 & 0.662 & 0.851 & 0.928 & 0.96
       \\
       
        & \cmark
       & 0.345 & 1.144 & 0.728 & 0.851 & 0.903
       \\
       
       \cmark & \cmark
      & \update{\textbf{0.095}}&	\update{\textbf{0.361}}&	\update{\textbf{0.909}}&	\update{\textbf{0.974}}&	\update{\textbf{0.986}}
       \\
      \hline
   \end{tabular}
  \end{center}
  \caption{Impact of different training losses. Here \emph{Temp.} refer to $L_{photo}^{off}$ (Eq.~\ref{eq:photo_temporal}) and \emph{Sparse} refers to $L_s$ (Eq.~\ref{eq:sparse_loss}).}
  \vspace{-0.3cm}
  \label{tab:ablation_losses}
\end{table}

\begin{table}[htp!]
  \begin{center}
  \resizebox{\linewidth}{!}{%
    \begin{tabular}{|c|c|c|c|c|c|}
      \hline 
       \emph{Sparsepoints (\%)}
       & Rel. $\downarrow$& RMSE $\downarrow$& $\delta_1$ $\uparrow$& $\delta_2$ $\uparrow$& $\delta_3$$\uparrow$
       \\
       \hline
       \hline

        100 \% & \textbf{0.095} & \textbf{0.361} & \textbf{0.909} & \textbf{0.974} & \textbf{0.986} \\
        75 \% & 0.113 & 0.521 & 0.866 & 0.950 & 0.973 \\
        50 \% & 0.115 & 0.526 & 0.862 & 0.950 & 0.972 \\
        25 \% & 0.119 & 0.537 & 0.852 & 0.947 & 0.972 \\
        0 \% & 0.136 & 0.581 & 0.803 & 0.932 & 0.969 \\
      \hline
   \end{tabular}
   }
  \end{center}
  \caption{$\%$ of sparse points used as input for inference.}
  \vspace{-0.5cm}
  \label{tab:ablation_sparsepoints}
\end{table}

Similarly, to understand the impact of different training losses we train multiple networks with different combinations of training losses and report results in Table~\ref{tab:ablation_losses}. Removing the sparse loss affects the performance most indicating its importance in providing supervision for far away regions where we do not have depth estimates. The temporal loss provides supervision on occluded regions together with temporal continuity and in combination, the two losses complement each other giving the best results.

\noindent\textbf{Number of sparse points.} Table \ref{tab:ablation_sparsepoints} shows the performance only drops slightly when number of sparse point is reduced. This means that \ACDC~can obtain accurate dense depth with as little as $60$ tracked landmarks. However, removing the sparse points completely leads to larger performance drop.

\begin{figure}[htp!]
     \centering
     \vspace{0.2cm}
     \includegraphics[width=\linewidth]{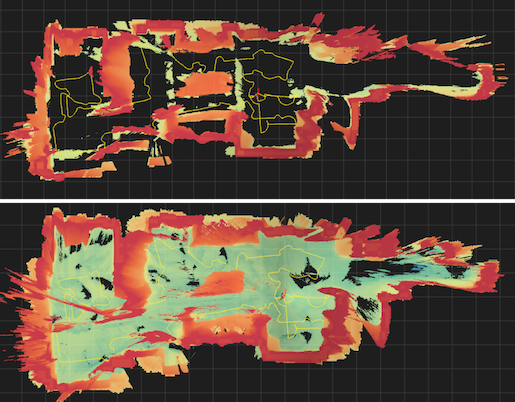}
     \caption{\textit{From top to bottom: 3D reconstructions from raw D435i and completed \ACDC-R50.}}
     \vspace{-0.5cm}
     \label{fig:results_reconstruction}
\end{figure}

\noindent\textbf{3D Mapping for Robotics.} Finally, we show how accurate depth completion can boost the performance of a down-stream task such as 3D mapping by recording a dataset using a wheeled robot 
with a setup similar to existing robotic vacuum cleaners. The reduced ground clearance of the camera results in significant portions of the floor missing from the depth map. This leads to a floor-less reconstruction, as seen in Fig.~\ref{fig:results_reconstruction}. \ACDC-R50~is able to complete the floor and other regions, significantly improving the map.

\section{Conclusions}\label{sec:conclusions}
In this work we investigate self-supervised depth completion for active stereo systems. We closely integrate a visual-inertial SLAM system into our training and inference pipelines, providing reliable pose estimates during training as well as accurate 3D landmarks. We propose \ACDC~that leverages these 3D landmarks as input and as weak supervision, resulting in more reliable depth estimates for distant areas. We further propose a novel reconstruction loss that relies on both passive and active frames, giving a valuable supervision signal in texture-less areas. We show through several ablation studies that all these losses and inputs contribute to the final result.
In addition, we contribute two datasets (synthetic and real) for active stereo depth completion and prediction. We believe this to be an important contribution due to the non-existence of active stereo datasets with ground truth in the community, and we hope that this will attract further research on this topic.



{\small
\bibliographystyle{ieee_fullname}
\bibliography{root}
}

\newpage

\section{Appendix}

The appendix contains more details about the channel exchanging network and the proposed $\max$ channel exchange mechanism. We further present explicit expressions for all the photometric losses used for training. We provide additional qualitative results and qualitatively show how different inputs contribute to the final loss. Lastly, we provide more details into the proposed datasets. 

\subsection{Channel Exchange Network}\label{sec:model_appendix}

To make the paper self contained, we provide some extra details into the channel exchanging network (CEN). We further highlight that taking the maximum across channels during the exchange, rather than the average as proposed in \cite{Wang20arxiv}, gives superior results. As mentioned in the main paper, we have three identical sub-networks, $f_m$, with shared weights (except for the Batch-Normalisation (BN) layers).
The final output $\hat{D}$ is a linear combination of the sub-networks with a decision score $\alpha_m$ learned with an associated softmax.

\begin{equation}
\hat{D} = \sum_{m=1}^{M} \alpha_{m} f_{m}\left(\boldsymbol{x}_{m}^{(i)}\right),
\end{equation}

where the input $\boldsymbol{x}^i$ is the initial depth, guidance image or sparse 3D landmarks. Channel exchange is achieved via the BN layers, where the individual channel importance that is measured by the magnitude of BN scaling factor $\gamma_{m, l}$ during training. \cite{Wang20arxiv} propose that a channel is replaced with the average across the other sub-networks, if the magnitude of the sub-network-specific BN scaling factor $\gamma_{m,l}$ is below a fixed threshold

\begin{equation}
\resizebox{0.9\linewidth}{!}{$
    \cramped{
        \boldsymbol{x}_{m, l, c}^{\prime}=
    \begin{dcases}
        \gamma_{m, l, c} \frac{\boldsymbol{x}_{m, l, c}-\mu_{m, l, c}}{\sqrt{\sigma_{m, l, c}^{2}+\epsilon}}+\beta_{m, l, c}& \text{if } \gamma_{m, l, c}>\theta\\
        \frac{1}{M-1}\sum^M_{m^{\prime} \neq m} \gamma_{m^{\prime}, l, c} \frac{\boldsymbol{x}_{m^{\prime}, l, c}-\mu_{m^{\prime}, l, c}}{\sqrt{\sigma_{m^{\prime}, l, c}^{2}+\epsilon}}+\beta_{m^{\prime}, l, c}              & \text{else}
    \end{dcases}
    }
$}
\end{equation}

In contrast to \cite{Wang20arxiv}, that replaces with average channel signal across the sub-networks, we propose to replace with the channel with the strongest signal, \ie  $\max$ over the sub-networks. 

\begin{equation}
    \resizebox{0.9\linewidth}{!}{$
    \cramped{
        \boldsymbol{x}_{m, l, c}^{\prime}=
    \begin{dcases}
        \gamma_{m, l, c} \frac{\boldsymbol{x}_{m, l, c}-\mu_{m, l, c}}{\sqrt{\sigma_{m, l, c}^{2}+\epsilon}}+\beta_{m, l, c}& \text{if } \gamma_{m, l, c}>\theta\\
        \max_{m^{\prime} \neq m} \gamma_{m^{\prime}, l, c} \frac{\boldsymbol{x}_{m^{\prime}, l, c}-\mu_{m^{\prime}, l, c}}{\sqrt{\sigma_{m^{\prime}, l, c}^{2}+\epsilon}}+\beta_{m^{\prime}, l, c}              & \text{else}
    \end{dcases}
    }
$}
\end{equation}

where $\mu_{m,l,c}$, $\sigma_{m,l,c}^2$, $\gamma_{m,l,c}$ and $\beta_{m,l,c}$ are the mean and variance for the batch, the learnable scaling and bias parameters for the $c^{th}$ channel in the $l^{th}$ layer in the $m^{th}$ sub-networks.

\begin{figure}
     \centering
     \begin{subfigure}[b]{0.49\linewidth}
         \centering
         \includegraphics[width=\textwidth]{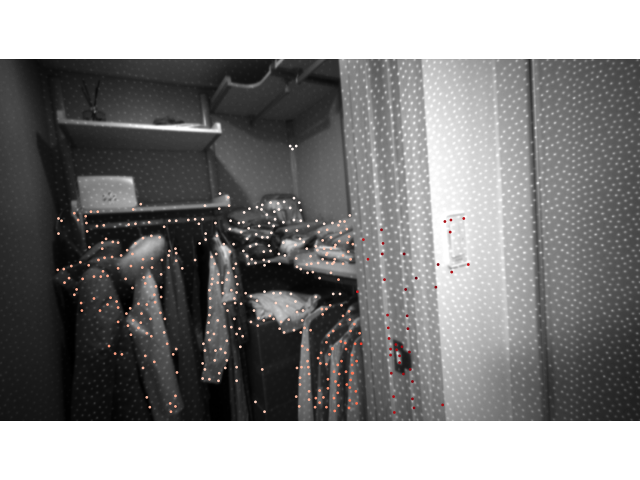}
     \end{subfigure}
     \begin{subfigure}[b]{0.49\linewidth}
         \centering
         \includegraphics[width=\textwidth]{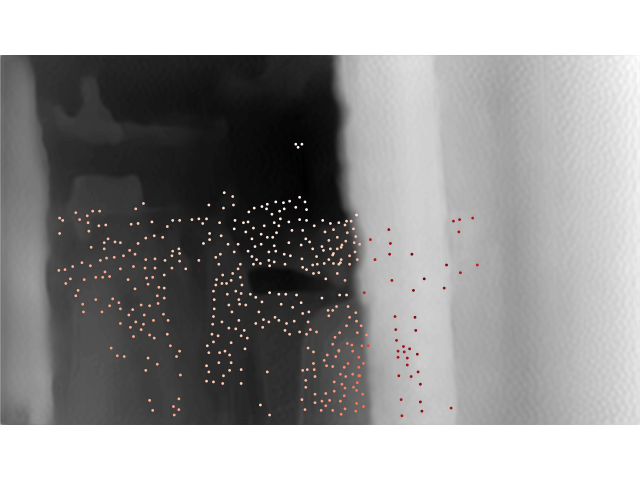}
     \end{subfigure}
     \begin{subfigure}[b]{0.24\linewidth}
         \centering
         \includegraphics[width=\textwidth]{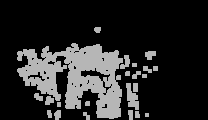}
         \caption{}
     \end{subfigure}
     \begin{subfigure}[b]{0.24\linewidth}
         \centering
         \includegraphics[width=\textwidth]{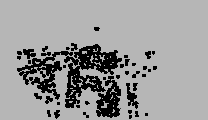}
         \caption{}
     \end{subfigure}
    \begin{subfigure}[b]{0.24\linewidth}
         \centering
         \includegraphics[width=\textwidth]{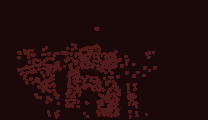}
         \caption{}
     \end{subfigure}
     \begin{subfigure}[b]{0.24\linewidth}
         \centering
         \includegraphics[width=\textwidth]{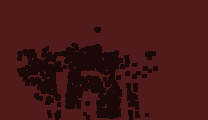}
         \caption{}
     \end{subfigure}
     \caption{\textit{Top row: Input infrared and predicted disparity, input sparse points are projected to both. Bottom row: Exchanged features for first ResNet block: (a) Disparity branch, feature \#0, (b) Disparity branch, feature \#19, (c) Infrared branch, feature \#0, (d) Infrared branch, feature \#19. Black, gray, red: features copied from sparse, disparity and infrared respectively.} Features of the sparse branch are copied to both disparity and infrared branches. The proposed 'maximum exchange' allows the network to route features from sparse directly, instead of an average of other branches.}
     \label{fig:exchange_max}
     \vspace{-0.4cm}
\end{figure}

In Fig. \ref{fig:exchange_max}, we show the channel exchange for a frame of D435i test set. We can observe that our method learns to route features from the sparse branch to disparity and infrared branches. 
Our proposal produces pixel-wise routing (see Fig. \ref{fig:exchange_max}), while the channel exchange proposed by \cite{Wang20arxiv} can not route from different branches, because they use an average of the other branches.

\subsection{Explicit Expression for Photometric Losses}\label{sec:losses_supplementary}
We use a combination of passive and active photometric losses. We propose photometric losses that operates on the active stereo images ($I_{t,L}^{on}$ and $I_{t,R}^{on}$) and the passive stereo images at the next time step ($I_{t+1,L}^{off}$ and $I_{t+1,R}^{off}$) and the previous time step ($I_{t-1,L}^{off}$ and $I_{t-1,R}^{off}$). We compute the photometric reprojection error between the stereo frames with the projector on $L_{stereo}^{on}$

\begin{equation}
    L_{stereo}^{on} = L_p(I_{t, L}^{on}, I_{t, R\rightarrow L}^{on}).
\end{equation}

We incorporate temporal consistency and a wider baseline by computing the photometric losses between the previous and next passive frames. This is achieved by reprojecting  both the previous and next passive images into the current view at time $t$. These passive images satisfy the photometric consistency constraint as neither have the active pattern. We compute the temporal losses, $L_{temp, R}^{off}$ and $L_{temp, L}^{off}$, for both the left/right stereo frame as:

\begin{align}
    L_{temp, R}^{off} &= L{p}(I_{t-1 \rightarrow t, R \rightarrow L}^{off}, I_{t+1 \rightarrow t, R \rightarrow L}^{off}), \\
    L_{temp, L}^{off} &= L{p}(I_{t-1 \rightarrow t, L}^{off}, I_{t+1 \rightarrow t, L }^{off}),
\end{align}
note that both left and right frames are projected into the current view which is in the left frame.

Lastly, since we have reprojected the previous/next left and right frames into the current frame, we can with minimal additional cost compute the passive stereo losses, $L_{stereo, t-1}^{off}$ and $L_{stereo, t+1}^{off}$, for the next and previous frames 

\begin{align}
    L_{stereo, t-1}^{off} &= L{p}(I_{t-1 \rightarrow t, L}^{off}, I_{t-1  \rightarrow t, R\rightarrow L}^{off}), \\
    L_{stereo, t+1}^{off} &= L{p}(I_{t+1  \rightarrow t, L}^{off}, I_{t+1  \rightarrow t, R\rightarrow L}^{off}).
\end{align}


\begin{figure*}[hpt!]
     \centering
     \begin{subfigure}[b]{0.19\linewidth}
         \centering
         \includegraphics[width=\textwidth]{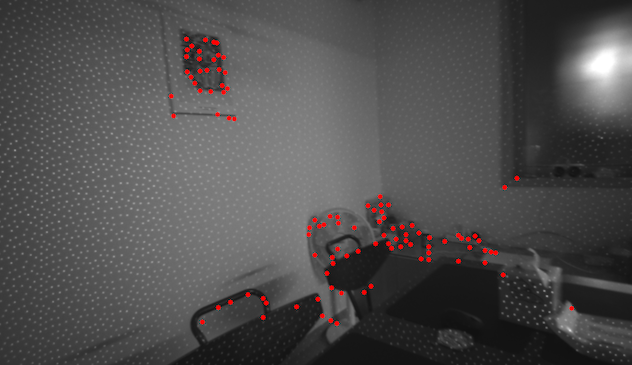}
     \end{subfigure}
     \begin{subfigure}[b]{0.19\linewidth}
         \centering
         \includegraphics[width=\textwidth]{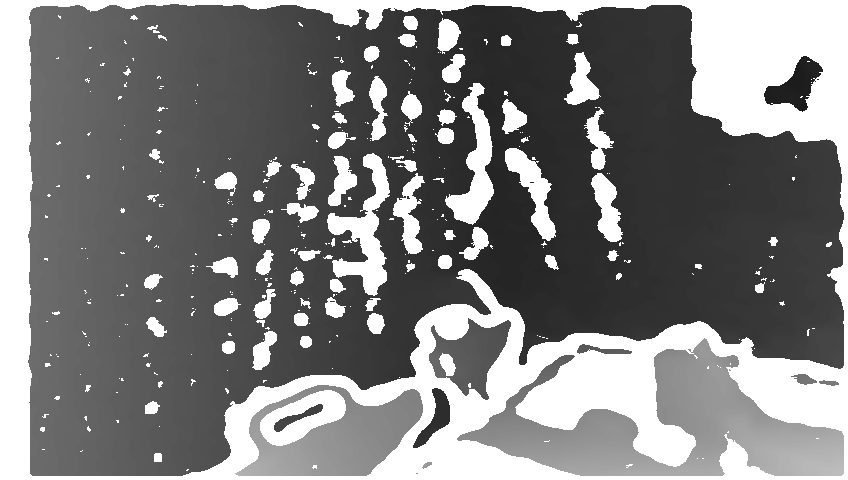}
     \end{subfigure}
     \begin{subfigure}[b]{0.19\linewidth}
         \centering
         \includegraphics[width=\textwidth]{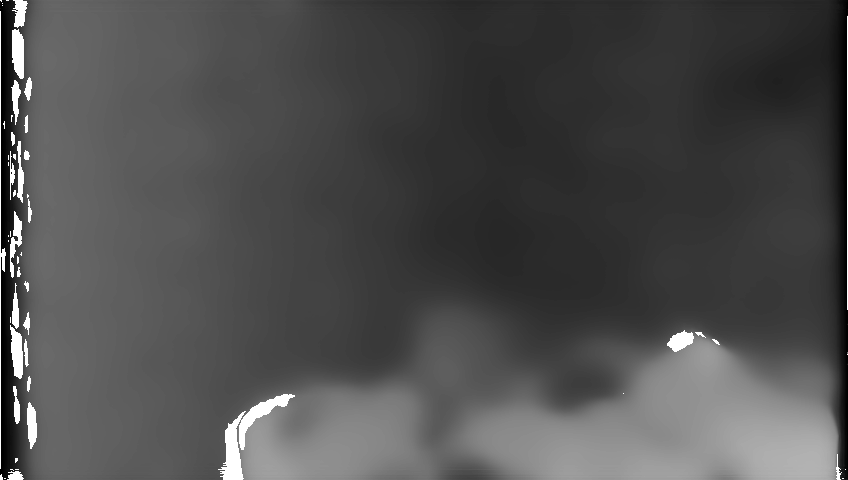}
     \end{subfigure}
     \begin{subfigure}[b]{0.19\linewidth}
         \centering
         \includegraphics[width=\textwidth]{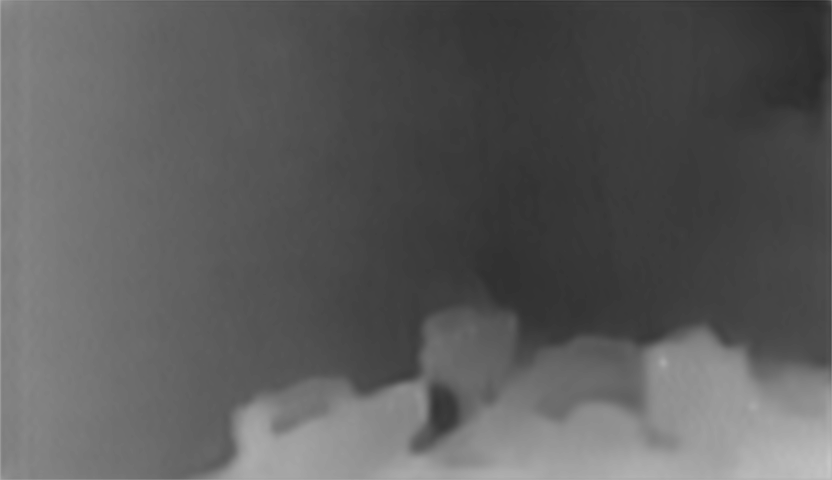}
     \end{subfigure}
    \begin{subfigure}[b]{0.19\linewidth}
         \centering
         \includegraphics[width=\textwidth]{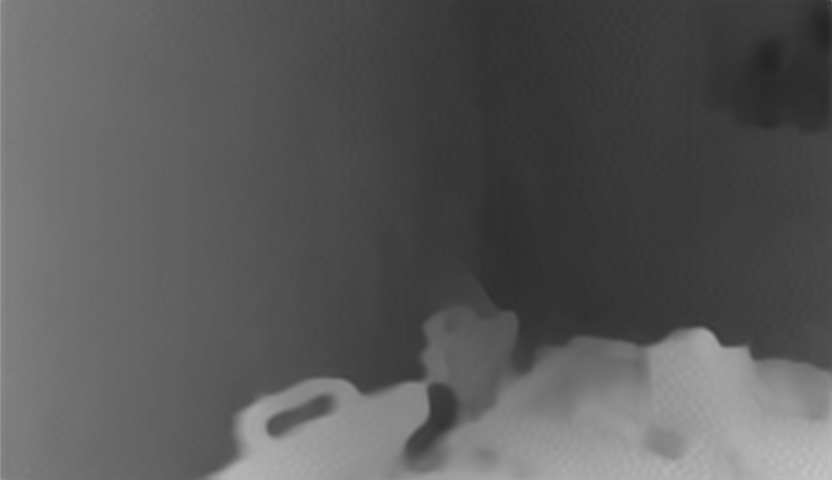}
     \end{subfigure}
     \begin{subfigure}[b]{0.19\linewidth}
         \centering
         \includegraphics[width=\textwidth]{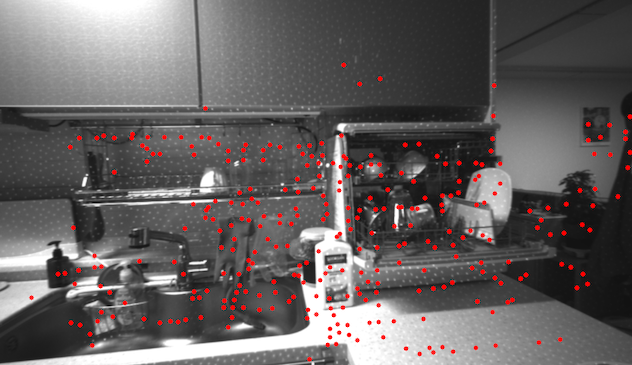}
     \end{subfigure}
     \begin{subfigure}[b]{0.19\linewidth}
         \centering
         \includegraphics[width=\textwidth]{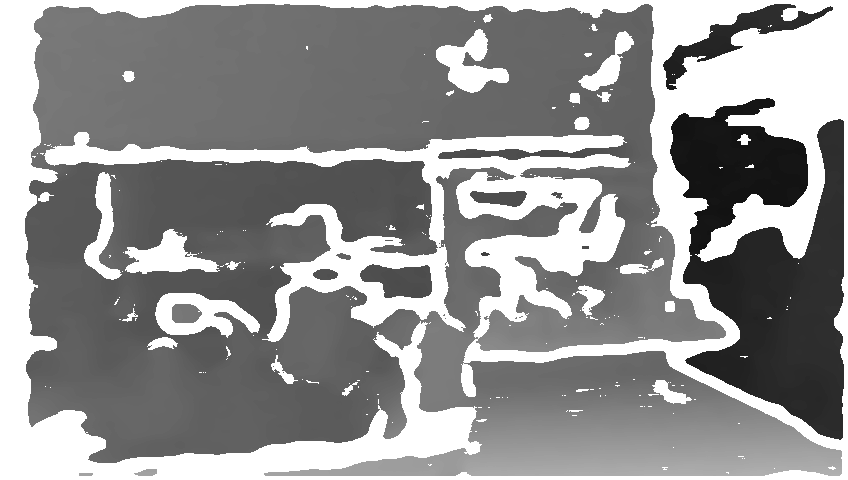}
     \end{subfigure}
     \begin{subfigure}[b]{0.19\linewidth}
         \centering
         \includegraphics[width=\textwidth]{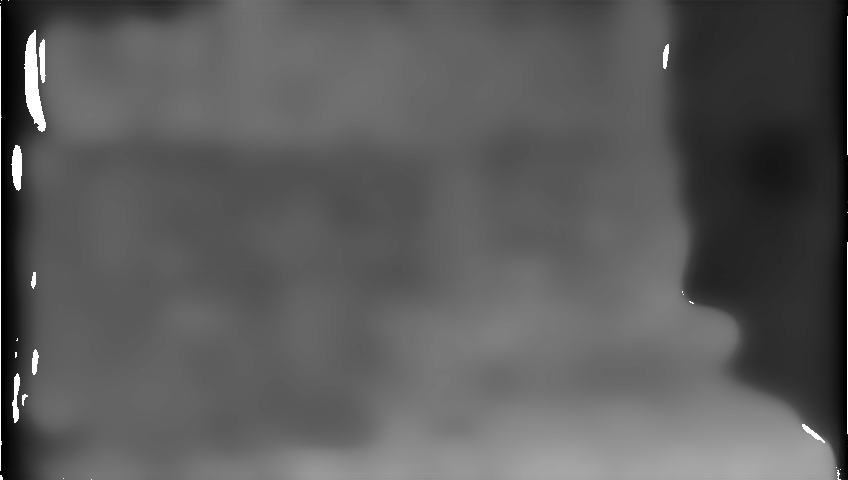}
     \end{subfigure}
     \begin{subfigure}[b]{0.19\linewidth}
         \centering
         \includegraphics[width=\textwidth]{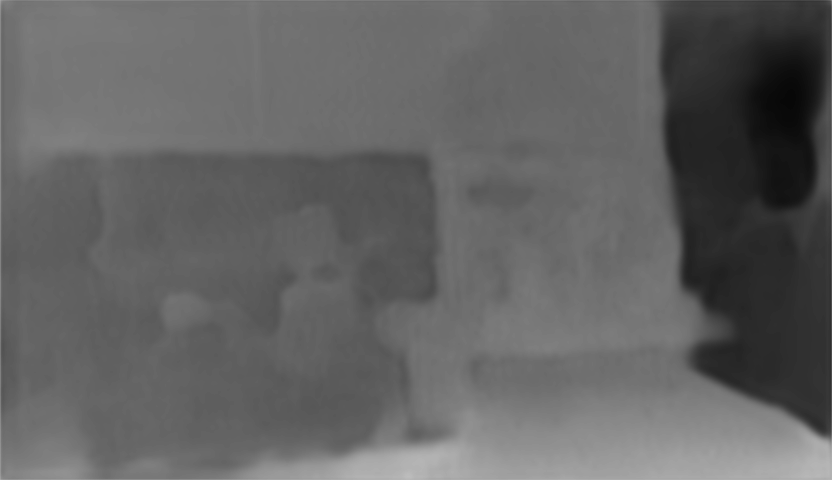}
     \end{subfigure}
    \begin{subfigure}[b]{0.19\linewidth}
         \centering
         \includegraphics[width=\textwidth]{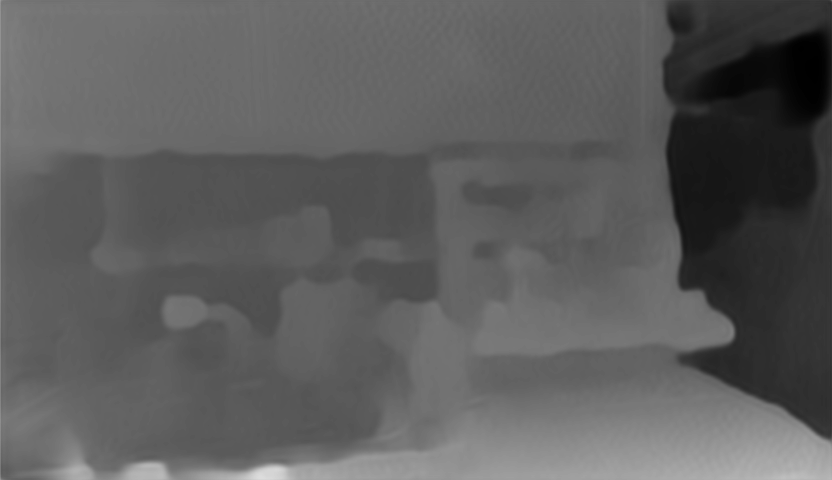}
     \end{subfigure}
     \begin{subfigure}[b]{0.19\linewidth}
         \centering
         \includegraphics[width=\textwidth]{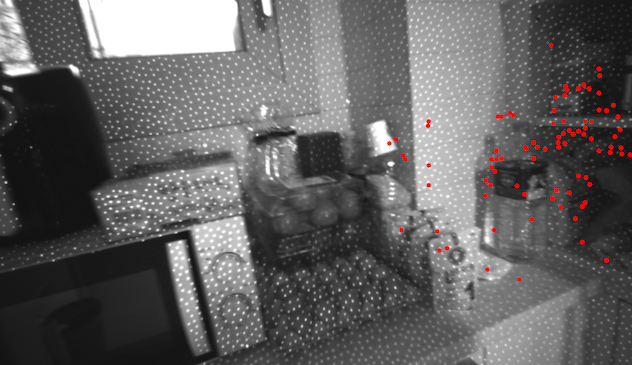}
     \end{subfigure}
     \begin{subfigure}[b]{0.19\linewidth}
         \centering
         \includegraphics[width=\textwidth]{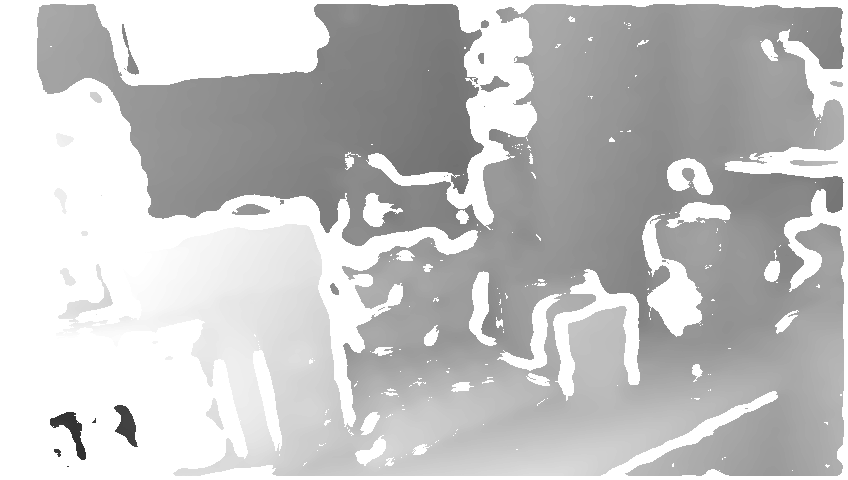}
     \end{subfigure}
     \begin{subfigure}[b]{0.19\linewidth}
         \centering
         \includegraphics[width=\textwidth]{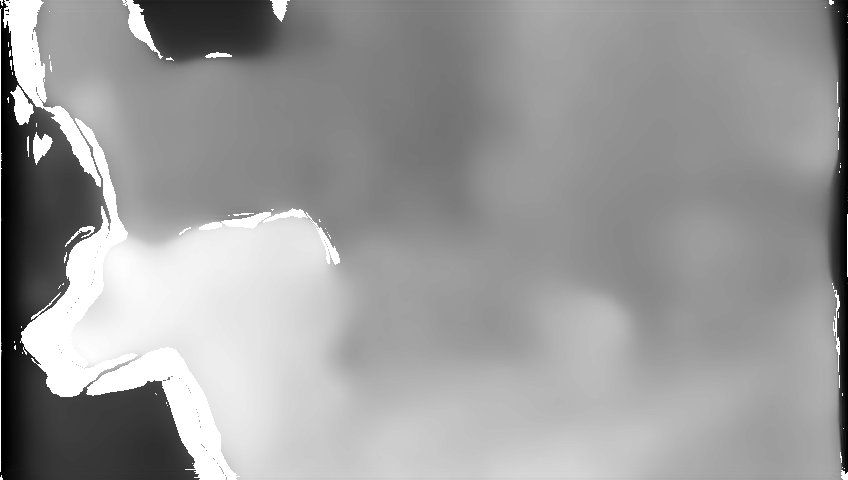}
     \end{subfigure}
     \begin{subfigure}[b]{0.19\linewidth}
         \centering
         \includegraphics[width=\textwidth]{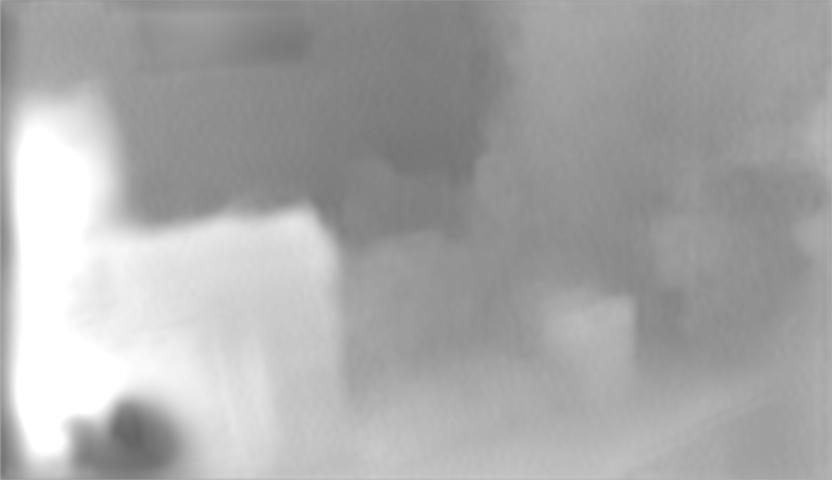}
     \end{subfigure}
    \begin{subfigure}[b]{0.19\linewidth}
         \centering
         \includegraphics[width=\textwidth]{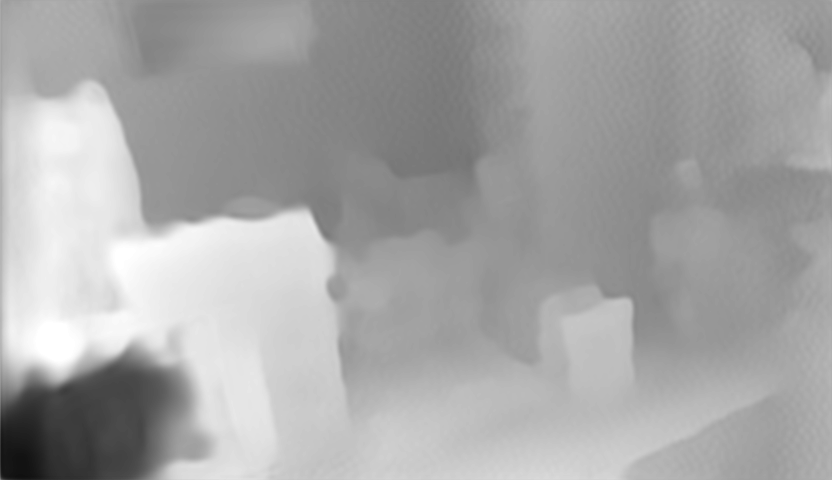}
     \end{subfigure}
     \begin{subfigure}[b]{0.19\linewidth}
         \centering
         \includegraphics[width=\textwidth]{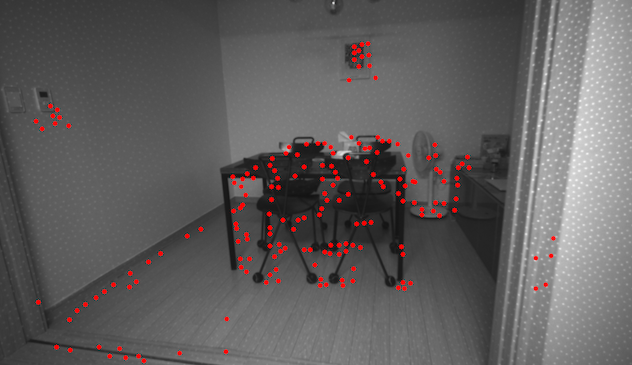}
     \end{subfigure}
     \begin{subfigure}[b]{0.19\linewidth}
         \centering
         \includegraphics[width=\textwidth]{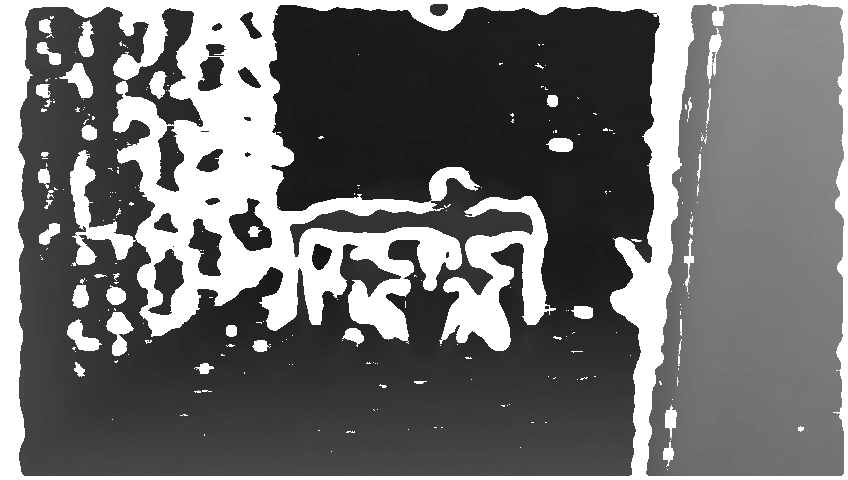}
     \end{subfigure}
     \begin{subfigure}[b]{0.19\linewidth}
         \centering
         \includegraphics[width=\textwidth]{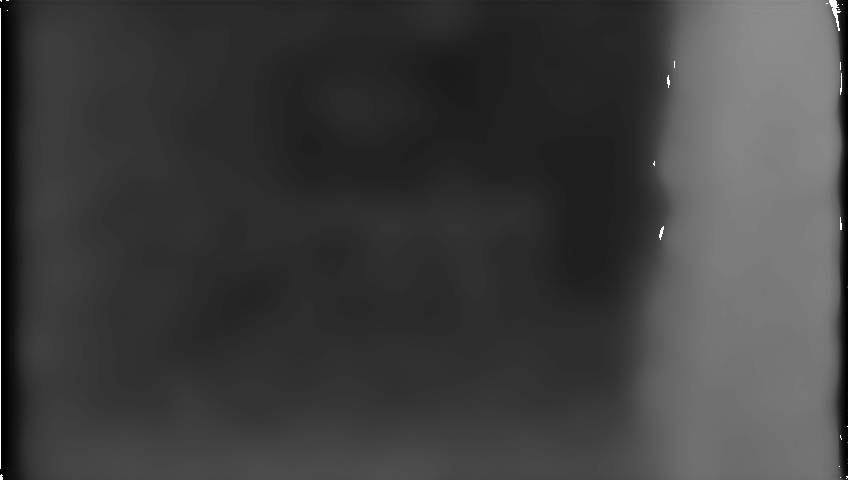}
     \end{subfigure}
     \begin{subfigure}[b]{0.19\linewidth}
         \centering
         \includegraphics[width=\textwidth]{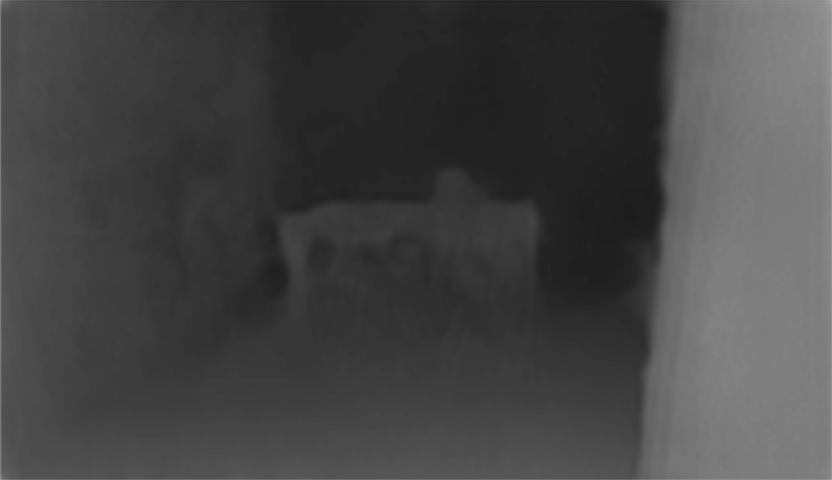}
     \end{subfigure}
    \begin{subfigure}[b]{0.19\linewidth}
         \centering
         \includegraphics[width=\textwidth]{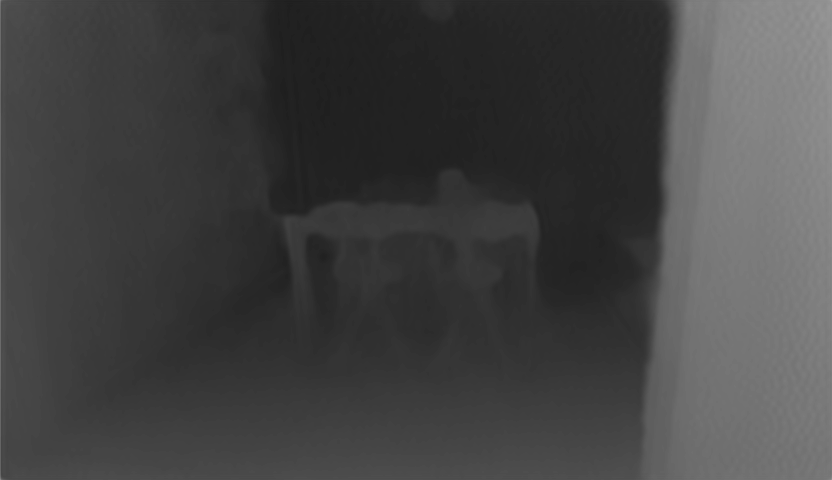}
     \end{subfigure}
     \begin{subfigure}[b]{0.19\linewidth}
         \centering
         \includegraphics[width=\textwidth]{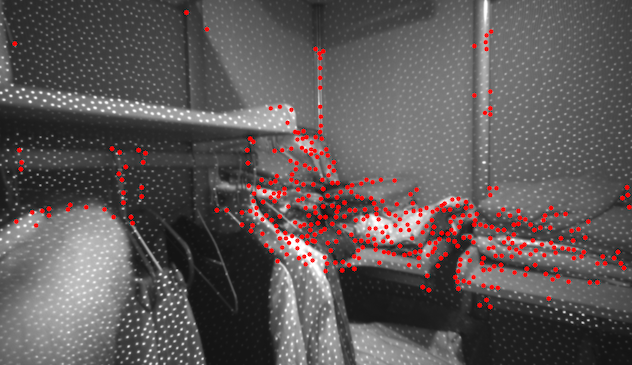}
     \end{subfigure}
     \begin{subfigure}[b]{0.19\linewidth}
         \centering
         \includegraphics[width=\textwidth]{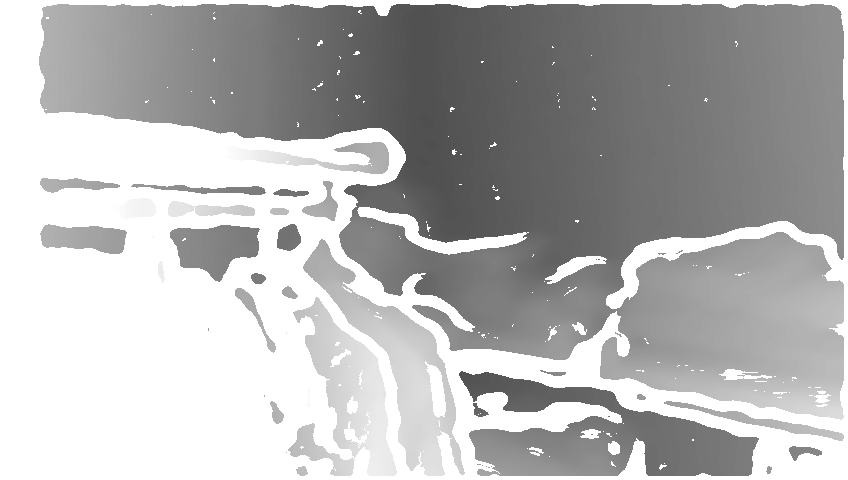}
     \end{subfigure}
     \begin{subfigure}[b]{0.19\linewidth}
         \centering
         \includegraphics[width=\textwidth]{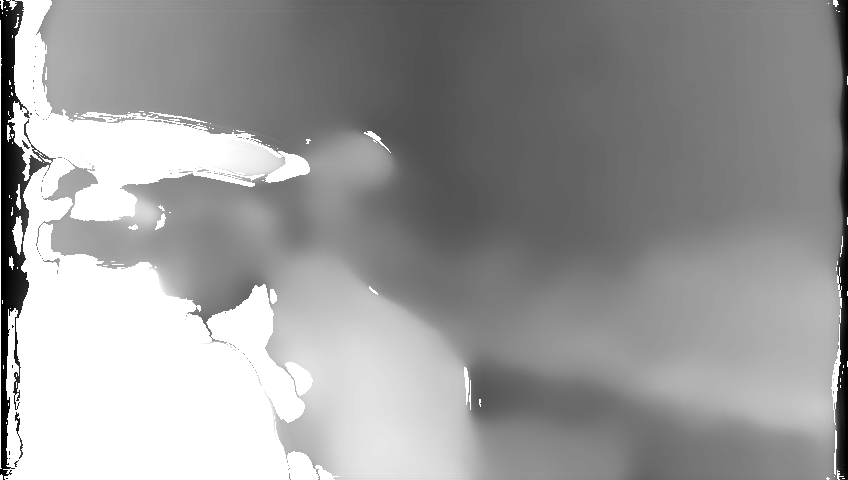}
     \end{subfigure}
     \begin{subfigure}[b]{0.19\linewidth}
         \centering
         \includegraphics[width=\textwidth]{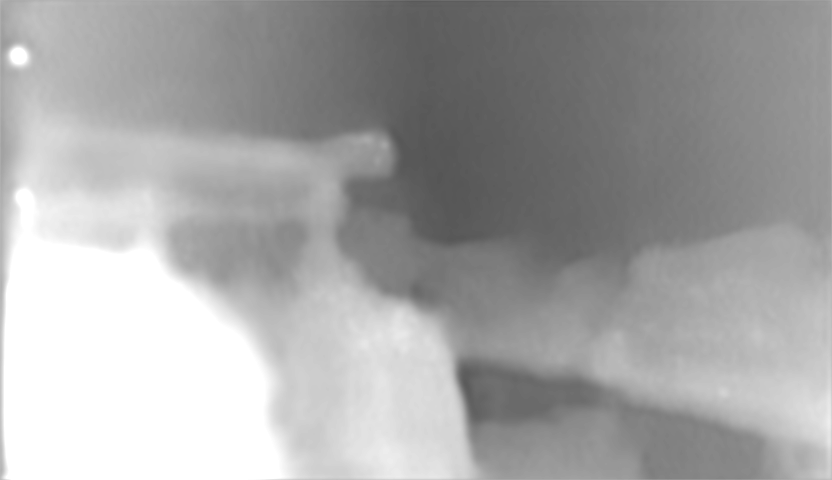}
     \end{subfigure}
    \begin{subfigure}[b]{0.19\linewidth}
         \centering
         \includegraphics[width=\textwidth]{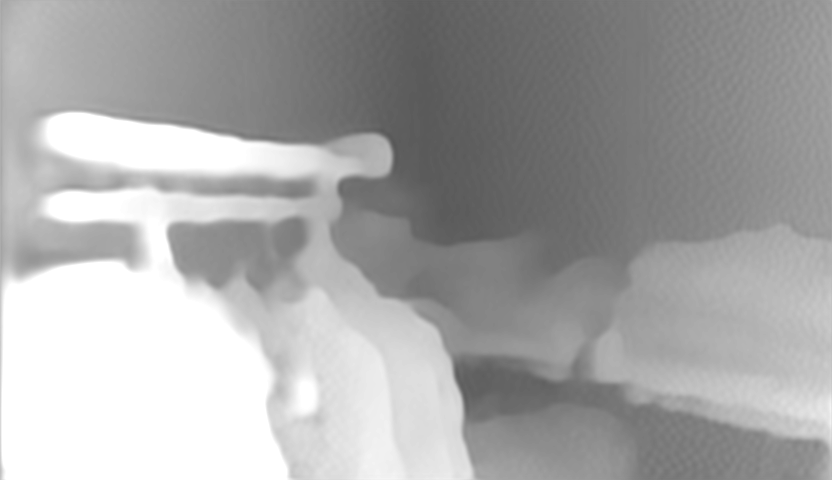}
     \end{subfigure}
     \begin{subfigure}[b]{0.19\linewidth}
         \centering
         \includegraphics[width=\textwidth]{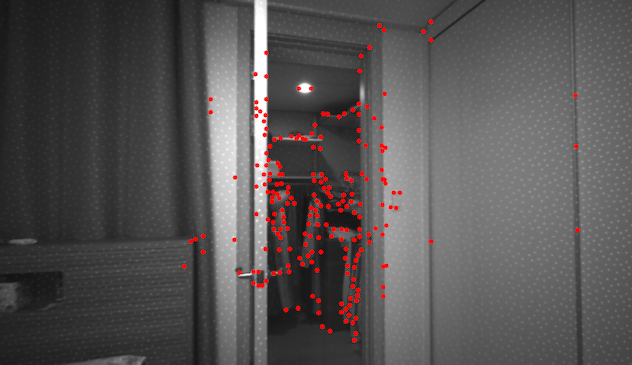}
     \end{subfigure}
     \begin{subfigure}[b]{0.19\linewidth}
         \centering
         \includegraphics[width=\textwidth]{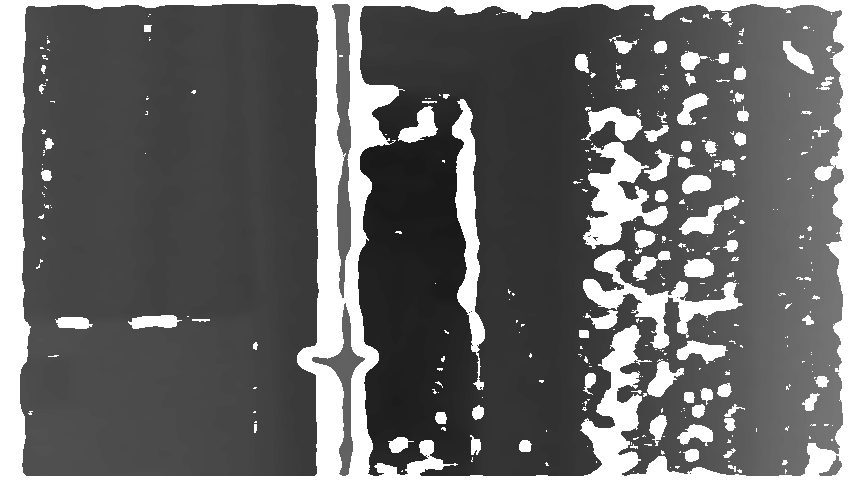}
     \end{subfigure}
     \begin{subfigure}[b]{0.19\linewidth}
         \centering
         \includegraphics[width=\textwidth]{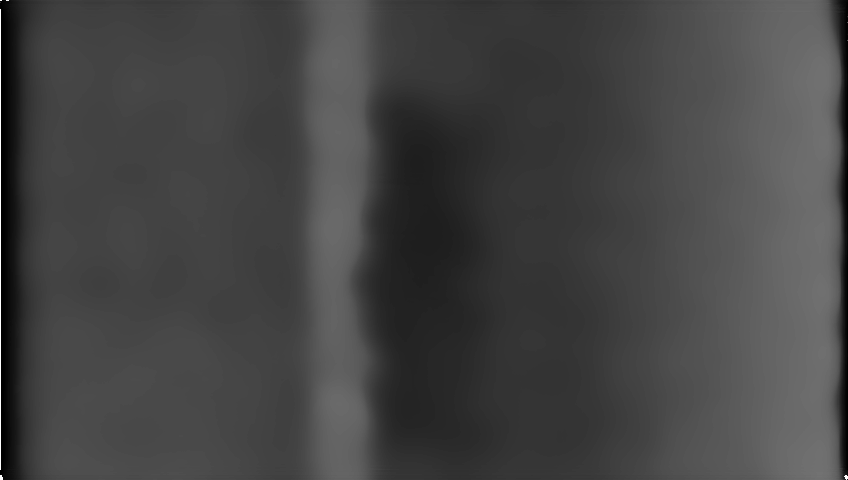}
     \end{subfigure}
     \begin{subfigure}[b]{0.19\linewidth}
         \centering
         \includegraphics[width=\textwidth]{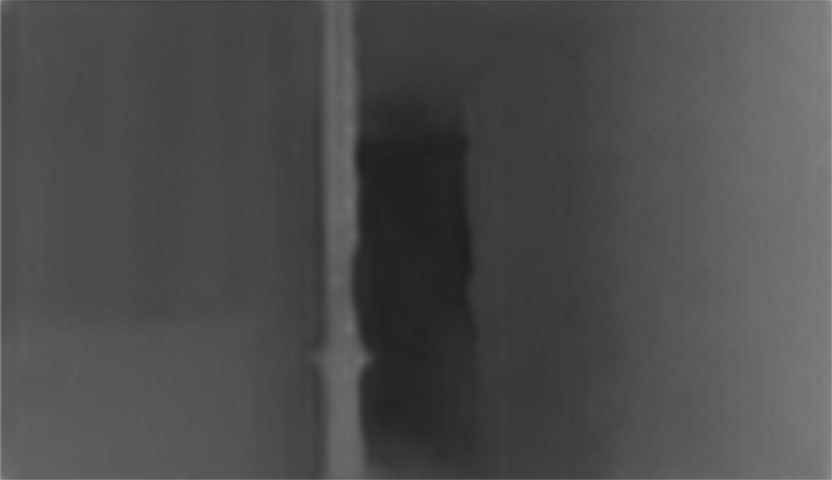}
     \end{subfigure}
    \begin{subfigure}[b]{0.19\linewidth}
         \centering
         \includegraphics[width=\textwidth]{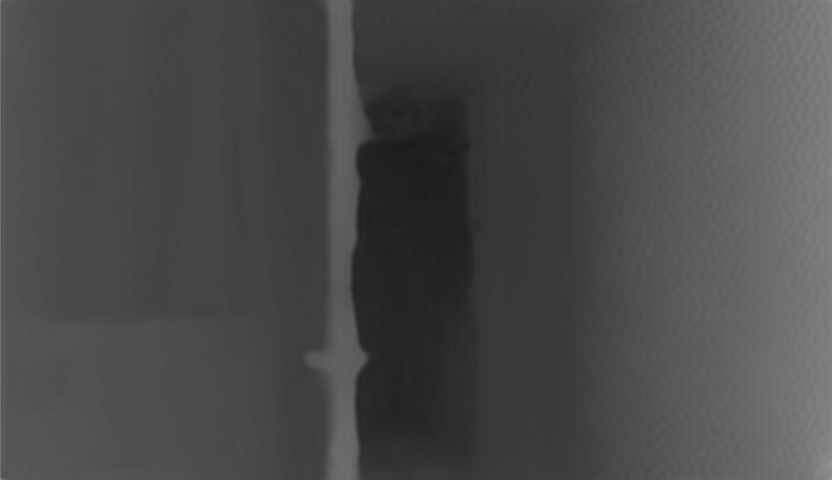}
     \end{subfigure}
     \begin{subfigure}[b]{0.19\linewidth}
         \centering
         \includegraphics[width=\textwidth]{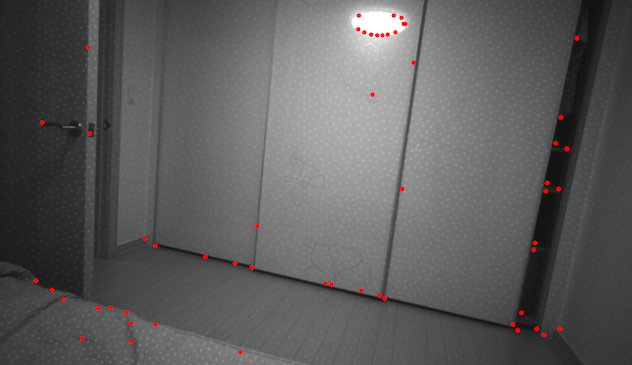}
     \end{subfigure}
     \begin{subfigure}[b]{0.19\linewidth}
         \centering
         \includegraphics[width=\textwidth]{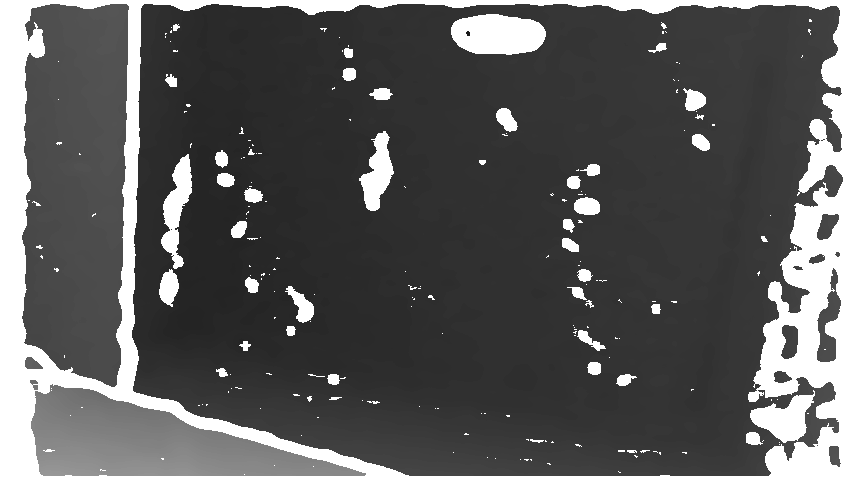}
     \end{subfigure}
     \begin{subfigure}[b]{0.19\linewidth}
         \centering
         \includegraphics[width=\textwidth]{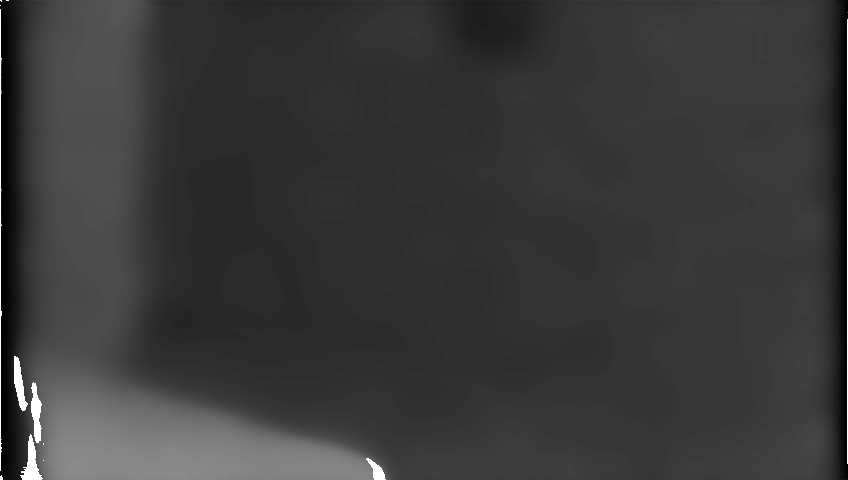}
     \end{subfigure}
     \begin{subfigure}[b]{0.19\linewidth}
         \centering
         \includegraphics[width=\textwidth]{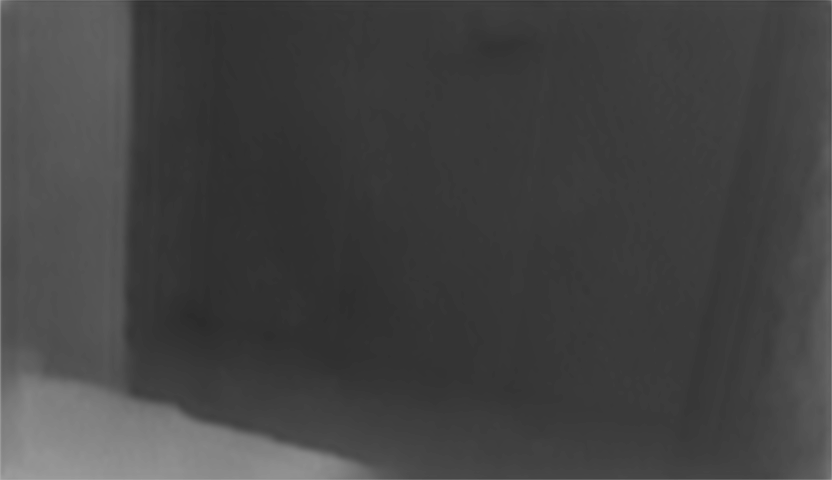}
     \end{subfigure}
    \begin{subfigure}[b]{0.19\linewidth}
         \centering
         \includegraphics[width=\textwidth]{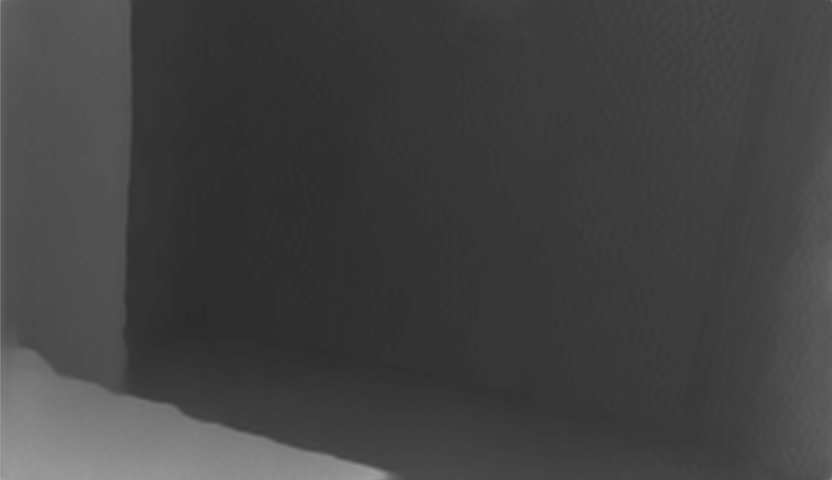}
    \end{subfigure}
     \begin{subfigure}[b]{0.19\linewidth}
         \centering
         \includegraphics[width=\textwidth]{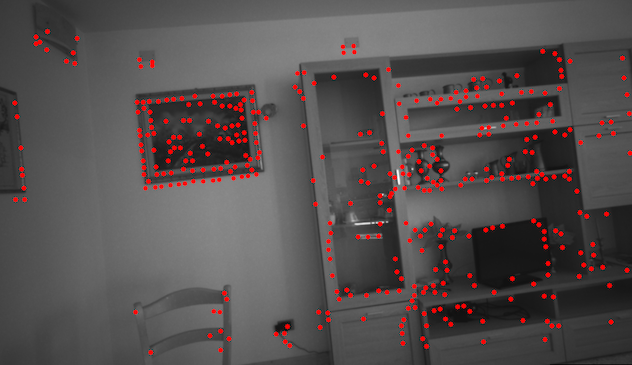}
     \end{subfigure}
     \begin{subfigure}[b]{0.19\linewidth}
         \centering
         \includegraphics[width=\textwidth]{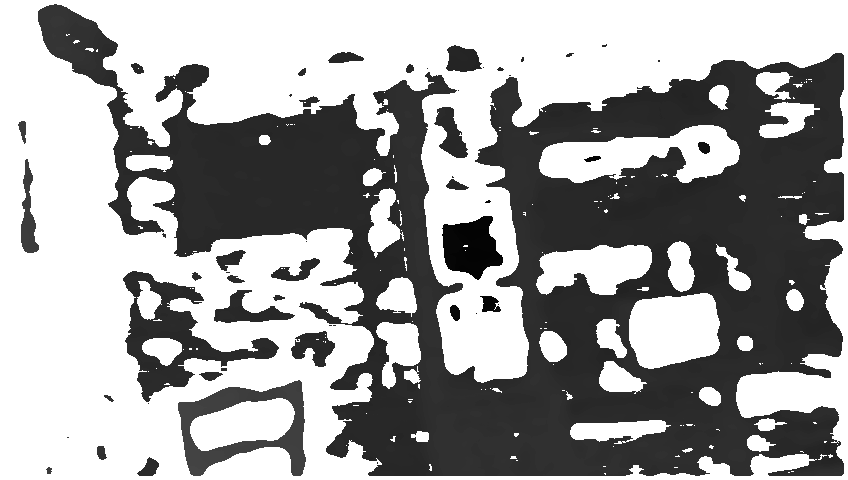}
     \end{subfigure}
     \begin{subfigure}[b]{0.19\linewidth}
         \centering
         \includegraphics[width=\textwidth]{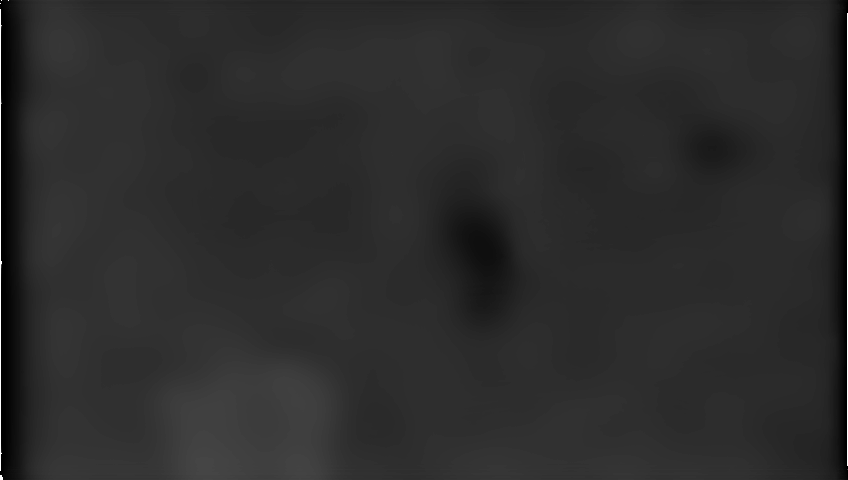}
     \end{subfigure}
     \begin{subfigure}[b]{0.19\linewidth}
         \centering
         \includegraphics[width=\textwidth]{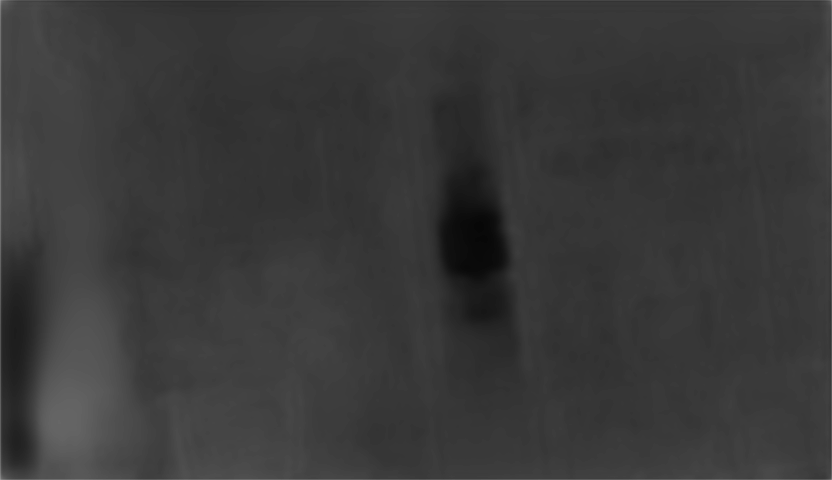}
     \end{subfigure}
    \begin{subfigure}[b]{0.19\linewidth}
         \centering
         \includegraphics[width=\textwidth]{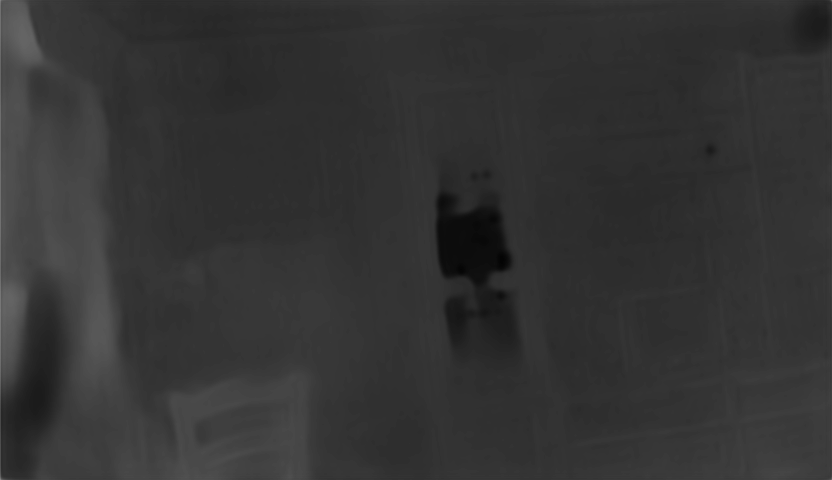}
     \end{subfigure}
     \caption{\textit{From left to right: Input infrared with projected sparse points, input disparity, \ASN~\cite{Zhang18eccv}, \ACDC-R18 and \ACDC-R50.} Qualitative comparison shows that our model produces accurate depth for challenging scenarios as thin structures (rows 1, 5 and 6), distant objects (row 4), lots of small objects (row 2), specularities (row 7). We also show two failure cases (rows 3 and 8).}
     \label{fig:results_realsense_supp}
\end{figure*}

\subsection{Additional qualitative results}
\label{sec:qualitative_results}

In Fig. \ref{fig:results_realsense_supp}, we provide additional qualitative results in the form of images from D435i test set. For each test sample, we show (a) infrared image with sparse points projected, (b) input disparity image, (c) \ASN~\cite{Zhang18eccv}, (d) \ACDC-R18 and (e) \ACDC-R50. We observe that our prediction is more complete than the input and best at conserving edges.

\begin{figure*}[hpt!]
\centering
     \includegraphics[width=\textwidth]{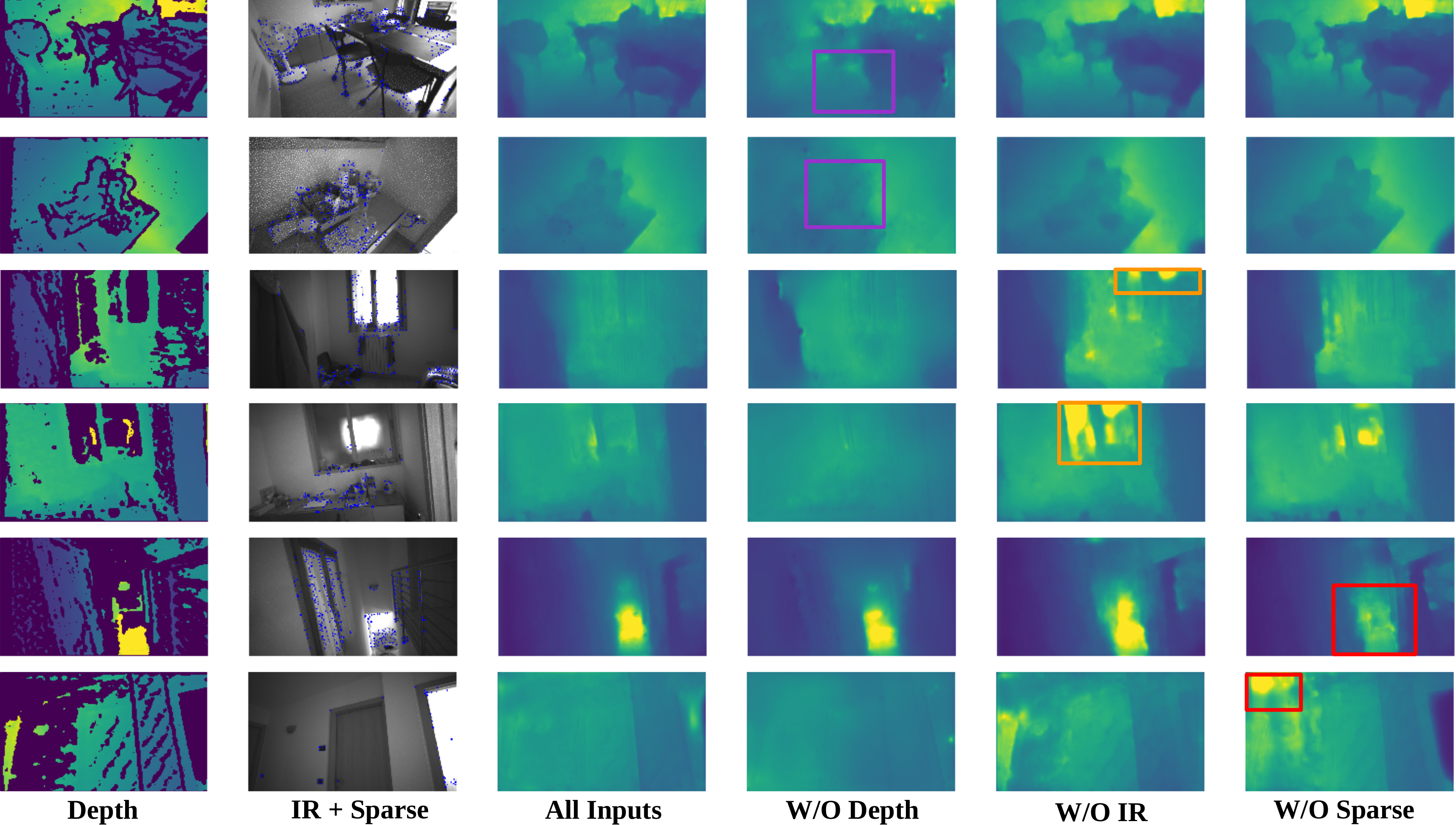}
     \caption{
     Here we compare the impact of different inputs by training multiple networks with different input combination. Purple boxes shows missed details due to unavailable input depth, orange boxes show missed context due to unavailable input IR guide image and red boxes show incorrect far away predictions due to unavailable sparse points.}
     \label{fig:results_input_comparison}
\end{figure*}

Fig. \ref{fig:results_input_comparison} compares the impact of different inputs for depth completion. We observe that while the incomplete depth input helps provide detail (purple regions), the guide IR encodes context to better complete the depth (orange regions). This is specially observed in cases of reflective or transparent surfaces. Finally, the sparse input helps anchor the depth completions in far away regions (red regions) where the input depth is often missing or has large noise.

\subsection{Dataset}\label{sec:dataset_supplementary}

As pointed out in the main paper, we did not find any publicly available dataset sufficient for our needs. Therefore, we presented two new datasets; Active TartanAir and D435i Sequences. In this section, we present a short review of existing datasets and give more details on the curation of Active TartanAir. 

\subsubsection{Related work}

\begin{table*}[htp!]
    \centering
    \begin{tabular}{c|c|c|c|c|c|c|c|c|c}
         Name & Synthetic & \begin{tabular}{@{}c@{}}(Pseudo) \\ GT depth\end{tabular}  & RGB or IR & 
         \begin{tabular}{@{}c@{}}Initial\\ depth maps\end{tabular} & GT poses & Stereo & IMU &
         \begin{tabular}{@{}c@{}}Active \\ texture\end{tabular}  & 
         \begin{tabular}{@{}c@{}}Interleave \\ mode\end{tabular} \\ \hline  \hline
         
NYU-Depth V2~\cite{Silberman:ECCV12}&&\checkmark&\checkmark&\checkmark&&&\checkmark&&\\

ScanNet~\cite{dai2017scannet}&&\checkmark&\checkmark&\checkmark&\checkmark& &\checkmark&  & \\
           
Matterport3D~\cite{Matterport3D}&&\checkmark&\checkmark&\checkmark&\checkmark&&&& \\
SceneNN~\cite{scenenn-3dv16}&&\checkmark&\checkmark&\checkmark&\checkmark&&&& \\

Stanford 
2D-3D-S~\cite{armeni2017joint}&&\checkmark&\checkmark&\checkmark&\checkmark&&&& \\
           
\begin{tabular}{@{}c@{}}
KITTI-\\ 
Depth Compl.~\cite{Uhrig17ic3dv}
\end{tabular}&&\checkmark&\checkmark&\checkmark&\checkmark&\checkmark&\checkmark&&\\

TUM RGB-D~\cite{sturm12iros}&&&\checkmark&\checkmark&\checkmark&&\checkmark&&\\

EuRoC~\cite{Burri25012016}&&\checkmark&\checkmark&&\checkmark&\checkmark&\checkmark&&\\

OpenLORIS~\cite{shi2019openlorisscene}&&&\checkmark&\checkmark&\checkmark&\checkmark&&\\

VOID~\cite{Wong20icra}&&&\checkmark&\checkmark&\checkmark&&\checkmark&&\\

ICL-NUIM~\cite{handaetalICRA2014}&\checkmark&\checkmark&\checkmark& &\checkmark&&&&\\

SceneNet RGB-D~\cite{McCormacetalICCV2017}&\checkmark&\checkmark&\checkmark& &\checkmark&&&&\\

Replica~\cite{replica19arxiv}&\checkmark&\checkmark$^*$&\checkmark$^*$& &\checkmark$^*$&\checkmark$^*$&&&\\

Hypersim~\cite{roberts2020}&\checkmark&\checkmark&\checkmark& &\checkmark&&&&\\

TartanAir~\cite{Wang20iros}&\checkmark&\checkmark&\checkmark& &\checkmark&\checkmark&&&\\

\hline
\textbf{D435i Seq.} &        &\checkmark&\checkmark&\checkmark&\checkmark&\checkmark&\checkmark&\checkmark&\checkmark \\
        
\textbf{ActiveTartanAir}&\checkmark&\checkmark&\checkmark&\checkmark&\checkmark&\checkmark&\checkmark&\checkmark&\checkmark\\ \hline
    \end{tabular}
    \caption{A comparative table for some of the currently available datasets with RGB-D or stereo sequences. The $^*$ mark denotes that the data is not provided, but can be generated with the provided SDK. Both synthetic and real datasets do not contain active textures and cannot be used in interleave mode. Existing real scene understanding datasets often miss some part of the sensor data required to run visual-inertial SLAM, while real SLAM datasets often lack structure ground truth. Synthetic datasets, on the other hand, lack some sensors and initial noisy or incomplete depth maps. Our two new datasets (Active TartanAir and D435i Sequences) are the first datasets designed for depth prediction and completion for active stereo sensors.}
    \label{tab:related_datasets}
\end{table*}

Table \ref{tab:related_datasets} summaries existing dataset often used for depth completion. Most of these datasets are not recorded with active stereo sensors, and therefore, do not have images with an active pattern projected into the scene. OpenLoris \cite{shi2019openlorisscene} is a dataset recorded with an Active Stereo sensor, however, it is not designed for depth completion and does not contain pseudo GT depth estimates. Due to the non-existence of active depth completion and prediction dataset for active stereo sensors, we present D435i Sequences and Active TartanAir (based on TartanAir~\cite{Wang20iros})  which contains GT depth, projected active texture and are recorded in interleave mode. The datasets will be made available upon acceptance.

\subsubsection{Active TartanAir}

We simulate an active stereo sensor by rendering a textured pattern into the scene using the ground truth depth. The pattern was obtained by analysing the real pattern projection on a white wall from the D435i sensor. This way we could detect blob positions by using a Difference of Gaussians (DoG) filter together with Non-Maximum Suppression. Recording in interleaved mode was also used to increase contrast by computing the difference between projector on and off. Additionally we model the variation of relative blob intensity with respect to distance experimentally by changing the distance of the sensor with respect to the wall. When doing the overlay on the synthetic images, for simplicity we assume that the projector is aligned to the left camera; we believe this also prevents the network learning predictions based on the position of the pattern in the guidance image, which may change from sensor to sensor. Occlusions in the right image are taken into account by considering the synthetic depth of both left and right images and requiring them to match.

\end{document}